\documentclass[11pt]{article}

\usepackage[suppress]{color-edits}
\usepackage{natbib}
\usepackage[algo2e,ruled,linesnumbered]{algorithm2e} % For algorithms

\usepackage{amsmath,amsfonts,amsthm}
\usepackage{xspace}
\usepackage[dvipsnames]{xcolor}
\usepackage{url}
\usepackage{authblk}
\usepackage{caption}
\usepackage{subcaption}
\usepackage{tikz}
\usepackage{paralist}
\usepackage{libertine}
\usepackage[
pagebackref,
colorlinks=true,
urlcolor=blue,
linkcolor=blue,
citecolor=OliveGreen,
]{hyperref}
\usepackage[nameinlink]{cleveref}
\usepackage{enumitem}

\addauthor{rdk}{blue}
\addauthor{pco}{purple}
\addauthor{vg}{PineGreen}
\addauthor{eg}{orange}

\newcommand{\expect}{\mathbb{E}}

\newcommand{\eps}{\varepsilon}

\newcommand{\RR}{\mathbb{R}} % Real numbers
\newcommand{\E}{\mathbb{E}}
\newcommand{\var}{\mathrm{Var}}
\newcommand{\prob}{\mathrm{P}}

\newcommand{\reals}{{\mathbb{R}}}

\def\dist{\textnormal{\texttt{dist}}}

\newcommand{\kldiv}[2]{{D_{KL} \left( #1 \, \| \, #2 \right)}}

\usepackage[margin=1in]{geometry}
\usepackage{txfonts}
\usepackage[T1]{fontenc}

\theoremstyle{definition}

\newtheorem{theorem}{Theorem}[section]
\newtheorem{lemma}[theorem]{Lemma}
\newtheorem{proposition}[theorem]{Proposition}

\title{Non-Stochastic CDF Estimation Using Threshold Queries}
\author[1]{Princewill Okoroafor}
\author[1]{Vaishnavi Gupta}
\author[1]{Robert Kleinberg}
\author[1]{Eleanor Goh}
\affil[1]{Cornell University, Ithaca, NY\protect\\
{\footnotesize \texttt{pco9@cornell.edu, vg222@cornell.edu, rdk@cs.cornell.edu, eg552@cornell.edu}}}
\date{}

\begin{document}

% Title page for title and abstract only.
\begin{titlepage}
\maketitle

\begin{abstract}
  \rdkedit{Estimating the empirical distribution of a scalar-valued 
  data set is a basic and fundamental task. In this
  paper, we tackle the problem of estimating an empirical
  distribution} in a setting with two challenging features. 
  First, the algorithm does not
  directly observe the data; instead, it only asks
  a limited number of threshold queries about each sample.
  Second, the data are not assumed to be independent
  and identically distributed;
  instead, we allow for an arbitrary process generating
  the samples, including an adaptive adversary. These
  considerations are relevant, for example, when modeling a
  seller experimenting with posted prices to estimate the
  distribution of consumers' willingness to pay for a product:
  offering a price and observing a consumer's purchase decision
  is equivalent to asking a single threshold query about their value,
  and the distribution of consumers' values may be non-stationary
  over time, as early adopters may differ markedly from late adopters.

  Our main result quantifies, to within a constant factor, the
  sample complexity of estimating the empirical CDF of a sequence
  of elements of $[n]$, up to $\eps$ additive error,
  using one threshold query per sample. The complexity depends only
  logarithmically on $n$, and our result can be interpreted
  as extending the existing logarithmic-complexity results for
  noisy binary search to the more challenging setting where
  noise is non-stochastic. Along the way to designing our
  algorithm, we consider a more general model in which the
  algorithm is allowed to make a limited number of
  simultaneous threshold queries on each sample. We
  solve this problem using Blackwell's Approachability Theorem
  and the exponential weights method.
  As a side result of independent interest, we
  characterize the minimum number of simultaneous
  threshold queries required by deterministic
  CDF estimation algorithms.
\end{abstract}

% SHORTENED ABSTRACT
% We present an algorithm to estimate the empirical distribution of a scalar-valued data set, when the data are adversarially generated and the algorithm may only ask a limited number of threshold queries about each data point rather than observing it directly. This estimation problem models, for example, a seller experimenting with posted prices to estimate the distribution of consumers' willingness to pay for a product: offering a price and observing a consumer's purchase decision is equivalent to asking a single threshold query about their value, and the distribution of consumers' values may be non-stationary over time, as early adopters may differ markedly from late adopters. Our main result quantifies, to within a constant factor, the sample complexity of estimating the empirical CDF of a sequence of elements of $[n]$, up to $\varepsilon$ additive error, using one threshold query per sample. The complexity depends only logarithmically on $n$, hence our result extends existing logarithmic-complexity results for noisy binary search to the more challenging setting where noise is non-stochastic. Along the way to designing our algorithm, we solve a more general problem with a limited number of simultaneous threshold queries per sample, using Blackwell's Approachability Theorem and the exponential weights method. As a side result of independent interest, we characterize the minimum number of simultaneous threshold queries required by deterministic CDF estimation algorithms.

\end{titlepage}

\section{Introduction}
\label{sec:intro}

\rdkedit{%
Estimating the empirical distribution of a scalar-valued 
data set is a basic and fundamental task. For example, 
estimating quantiles of a data stream is one of the
oldest and most well-studied problems in streaming algorithms
\citep{greenwald2001space,karnin2016optimal,manku1998approximate,munro1980selection}, with applications to databases~\citep{greenwald2001space}, network health monitoring~\citep{cormode2004holistic},
and wireless sensor networks \citep{shrivastava2004medians}, 
among others. Ideally, a data analyst would like to be able
to assume that the data values 
are independent and identically distributed,
and that they are directly observable.  
However, these assumptions might
be violated in applications of interest.}
\begin{enumerate}
\item In many settings, samples can only be evaluated indirectly
% using {\em threshold queries}, which reveal whether
% or not a sample is greater than or equal to a given threshold.
% This is the way a seller gains information
by comparing them to specified thresholds and
learning whether or not each sample is less
than or equal to its corresponding threshold.
This is the case, for instance, when a seller
experiments with varying posted prices in order to estimate
the distribution of consumers' willingness to pay for a
product or service. Other examples arise when
eliciting information about the distribution of
individuals' abilities using pass-fail tests
with a variable level of difficulty (e.g.~swimming tests)
or when evaluating the quality of a new product by asking
consumers to compare it against products of known quality.
(There is ample evidence in the behavioral sciences that
human subjects' quality judgments can be elicited more
reliably with ordinal comparisons than with subjective
numerical ratings
\citep{ali2012,chapelle,larichev,moshkovich}.)
\item The assumption that samples are
independent and identically distributed may also be violated. Returning to the posted-pricing
application, early adopters of a product might differ
markedly from late adopters in their willingness to pay
for the product, and the late adopters' willingness
to pay may even depend on the rate of adoption by earlier
consumers, which in turn depends on the posted prices
they were offered. \vgedit{A similar application arises in an auction setting with repeated bidding - an internet advertiser estimates the distribution of winning bids by varying their own bid. To account for complex influences on the behavior of other bidders, assuming a worst-case input sequence rather than i.i.d. is very useful \citep{weed16-auctions}.}
\end{enumerate}
In this work we tackle the problem of estimating
the empirical distribution of a sequence of numbers
using threshold queries, in a non-stochastic setting
that makes no assumptions about the process by which
the sequence is generated. Our model even allows the
sequence to be constructed by an adaptive adversary.
% as long as
% the adversary sees only the queries issued by the algorithm,
% not the random bits it will use to select future queries.
We assume the algorithm asks one threshold
query about each element of the sequence,
and the query and its answer are revealed
to both parties before the next element
of the sequence is generated by the adversary.
The key question we aim to resolve is: {\em what is the
sample complexity of estimating the empirical CDF of a
distribution on $[n] = \{1,2,\ldots,n\}$ to within $\eps$?}
In more detail, what is the smallest $T$ such that
there exists a randomized algorithm that succeeds,
with probability at least $3/4$, in learning an
estimate of the empirical CDF of
an arbitrary sequence $x_1,\ldots,x_T$
that differs from the true empirical CDF
(in $L_\infty$ norm) by at most $\eps$?
In this paper, we resolve the question to within
a constant factor, by proving asymptotically matching
upper and lower bounds. In fact,
our lower bound is valid even in a stochastic setting
where the elements $x_1,\ldots,x_T$ are i.i.d.~samples
from a distribution on $[n]$. Hence our results reveal,
perhaps surprisingly, that {\em up to a constant factor,
there is no difference in the sample complexity of
CDF estimation in the stochastic and non-stochastic
settings.}

\subsection{Relation to noisy binary search and median estimation}
\label{sec:noisy-binary}

Let us say that $m \in [n]$ is an {\em $\eps$-approximate
median} of the sequence $x_1,x_2,\ldots,x_T$ if at least
$(\frac12 - \eps) T$ elements of the sequence are
less than or equal to $m$ and at least $(\frac12 - \eps) T$
of them are greater than or equal to $m$.
Approximate median estimation reduces to
approximate CDF estimation: if $\hat{F}$ is an
$\eps$-accurate estimate of the empirical CDF of
$x_1,\ldots,x_T$ then an index $m$ that satisfies
$\hat{F}(m-1) < \frac12 \le \hat{F}(m)$
is an approximate median.

In the special case when $x_1,x_2,\ldots,x_T$
is restricted to be a constant sequence,
% (i.e.~$x_1 = x_2 = \cdots = x_T = x$ for
% some $x \in  [n]$)
CDF estimation
and median estimation both become equivalent to binary
search: the empirical CDF is a $\{0,1\}$-valued step
function with a step at some $x \in [n]$ and $x$ is
the unique approximate median, so both tasks become
equivalent to identifying the value of $x$ using queries
of the form $x \stackrel{?}{\le} q_t$. The problem
our work addresses can
thus be interpreted as a generalization of binary search
in which the answers to comparison queries are perturbed
by non-stochastic noise.

One easy consequence of this connection to binary search
is a lower bound of $\lfloor \log_2(n) \rfloor$
on the sample complexity of approximate CDF estimation and
approximate median estimation.
% \rdkcomment{Changed ceiling to floor function, because
% we allow error probability $\frac14$. For example, when
% $n = 2^k + 1$ and $x_1,\ldots,x_T$ is constant sequence
% there's an algorithm with sample complexity $k$ that
% initially guesses an element of $[n]$ at random and
% decides to ignore this element, then does binary
% search on all the other elements. This algorithm
% outputs the correct answer except when
% it is the random ignored element.}
In the important special
case when $\eps$ is a small constant (e.g., $\eps = 0.01$),
the algorithms we present in this paper match this trivial
lower bound to within a constant factor, exponentially
improving the best previously known bounds for CDF estimation
and median estimation in the non-stochastic setting.

\subsection{Techniques}

Given that CDF estimation generalizes binary search and
that the sample complexity bound we are aiming for ---
$O(\log n)$ in the case of constant $\eps$ --- matches
the query complexity of binary search, a natural
idea is to try designing CDF estimation algorithms
with a recursive structure resembling that of binary
search. Indeed, in the stochastic setting,
\citet{karp} presented a median estimation algorithm,
based on binary search with backtracking, whose sample
complexity is $O(\log n)$ when $\eps$ is constant.
Using this algorithm as a subroutine,
\citet{meister} showed how to solve CDF estimation
in the stochastic setting at the cost of an additional
$1/\eps$ factor in sample complexity.
In the non-stochastic setting, one can similarly attempt
to base CDF estimation or median estimation on divide-and-conquer
strategies that zero in on intervals where the
density of samples is high.
However, there is an obvious difficulty:
the past samples need not have any relation to
those in the present and future.
Thus, focusing on intervals that contained many
past samples could draw the algorithm's attention
away from the intervals containing most of the
present samples, making it impossible
to maintain an accurate CDF estimate.
We are not aware of any way to overcome
this difficulty and base a non-stochastic
CDF estimation algorithm on the principle
of divide-and-conquer with backtracking.

Instead, to design our algorithm we take a
detour through a more general model in which
the CDF estimation algorithm is allowed to
make $k$ simultaneous threshold queries for
each sample. When $k=1$ this matches our
original model, but when $k$ exceeds
$\frac{1}{\eps}$ the problem undergoes
an interesting qualitative change: it
becomes solvable by deterministic
algorithms. To solve it, we show that
the problem of using threshold queries
to compute a CDF estimate that is
accurate with probability 1 is equivalent
to a question about the approachability
of a convex set in a two-player game
with vector payoffs. Blackwell's
Approachability Theorem gives us a
criterion for determining the number
of simultaneous queries necessary to
solve CDF estimation using a Las Vegas
randomized algorithm that almost surely
terminates and outputs an $\eps$-accurate
answer. Using the exponential weight
approachability algorithm of
\citet{perchet}, we show that this
objective can in fact be achieved by
a deterministic algorithm in only
$O(\log(n)/\eps)$ rounds, with
$O(1/\eps)$ simultaneous queries
per round. We believe the design and analysis
of this deterministic, simultaneous-query
algorithm for CDF estimation may be
of independent interest. It is also a
vital step in designing a randomized
algorithm that solves CDF estimation in
the original non-stochastic setting with
only one threshold query per sample.
Our algorithm for that problem can be
interpreted as a randomized simulation
of the deterministic simultaneous-query
algorithm: it randomly samples one of the
$O(1/\eps)$ simultaneous queries recommended
by the deterministic algorithm, then uses
importance weighting to produce an unbiased
estimate of the payoff vector that would
have resulted from making all of the recommended
queries simultaneously.

\subsection{Related work}
\label{sec:relwork}

As noted earlier, our problem generalizes noisy
binary search to a setting with non-stochastic noise.
The first paper to study this generalization
is by \citet{meister}, who proved a sample
complexity upper bound $O(n \log(n) / \eps^2)$
using a na\"{i}ve algorithm that queries a
uniformly random threshold $q_t \in [n]$ at each time
$t \in [T]$ and estimates $\hat{F}(i)$ by simply
averaging the values observed in the time steps
$t$ when $q_t = i$. In other words, the na\"{i}ve
algorithm breaks down the problem of estimating a
CDF over $[n]$ into $n$ independent point-estimation
problems, one for each $i \in [n]$, which are each
solved by directly querying $F(i)$ often enough
that the average of the sampled queries approximates
the population average. This ignores the fact that
the empirical CDF must be a monotone function, and
that shape constraints such as monotonicity typically
improve the sample complexity of estimation \citep{isotonic}.
%\rdkcomment{Find a citation to a paper on the sample
%complexity of isotonic regression and insert it here.}
Perhaps surprisingly, \citeauthor{meister}
showed that when $\eps = O(1/n)$, there is a
lower bound for $\eps$-accurate CDF estimation
that matches the na\"{i}ve algorithm's sample
complexity up to a constant factor. This still
left an exponential gap between the upper and lower
bounds for the case of general $\eps > 0$.
Our work closes this gap, proving a tight bound
(up to constant factors) for all $n$ and $\eps > 0$
which exponentially improves
the Meister-Nietert upper bound in the case
$\eps = \Omega(1)$. Prior to our work, it was not known whether CDF estimation
algorithms could obtain any asymptotic improvement
at all over the na\"{i}ve algorithm.

The earliest works on noisy binary search
assumed a stochastic noise model that correctly
answers each comparison query with probability
$\frac12 + \eps$ and otherwise flips the answer.
In this model, an algorithm with sample complexity
$O(\log(n) / \eps^2)$ was presented and analyzed
by \citet{burnashev}, who actually
showed that their algorithm's complexity
is optimal up to a $1 + o(1)$ factor.
An even more precise sample complexity
bound for the same algorithm was later provided
by \citet{benor}. In the same model of stochastic noise,
\citet{feige} provided a different noisy binary
search algorithm with $O(\log(n)/\eps^2)$
complexity; they also supplied algorithms for
a number of other
fundamental problems such as sorting in the same
noisy comparison model. \citet{karp} generalized
the stochastic noisy comparison model to a setting
in which the probability of a correct answer to a
query depends on the identities of the elements
being compared, but is always greater than $\frac12$,
and they showed that the $O(\log(n)/\eps^2)$ sample
complexity bound for noisy binary search continues
to hold in this setting. \citet{meister} showed
how to use the Karp-Kleinberg noisy binary search
algorithm as a subroutine in a CDF estimation
algorithm that achieves sample complexity
$O(\log(n)/\eps^3)$ when the adversary is
stochastic. (In the notation introduced earlier,
this means the sequence $x_1,x_2,\ldots,x_T$ is
created by drawing i.i.d.~samples from a fixed
but unknown distribution.)

\egedit{Many works in the literature have studied distribution learning under specific constraints. \cite{han} and \cite{barnes} proved various minimax lower bounds for the problem of learning structured distributions in distributed networks. In their setting, every node in the network observes one independent sample drawn from the underlying distribution, and there is a central processor to which every node in the network communicates $k$ bits. \cite{acharya-information-constraints} significantly generalized \citeauthor{barnes}'s result, presenting unified lower bounds for distributed parametric estimation under a wide variety of local information constraints including communication, privacy, and data access constraints. They modeled their setting by considering a set of channels through which samples are passed to obtain observations. Using this general model, \citeauthor{acharya-information-constraints} were able to recover the bounds presented by \citeauthor{barnes}. A similar setting where $n$ nodes can observe $m$ samples and communicate information using $l$ bits was presented and analyzed in \cite{acharya-multiple-samples}. Like previous works, we model our problem as distribution learning under a set of constraints. Rather than focusing on communication channels in distributed models, our constraints are defined on the sequence of threshold queries the algorithm can make.} 

The distribution estimation problem is also studied within the context of online dynamic pricing and auctions. A seller repeatedly interacts with a buyer by setting prices for an item and observing whether the buyer purchases or not. An auctioneer learns a winning distribution by adaptively choosing reserve prices. Although both of these settings are limited to binary feedback: observing whether the item is bought, the literature \citep{kleinberg, leme, blum, reviewer1request} in these contexts assume the distribution is fixed or slowly-changing and often while trying to minimize a notion of regret with respect to the best fixed price in hindsight. We don't make any assumptions about the distribution.

% \rdkcomment{Section to be written later.
% For now, just some remarks on topics to
% cover.}
% \begin{enumerate}
% \item
% \rdkcomment{DONE!}
% Noisy binary search: Burnashev-Zigangirov,
% Feige-Peleg-Raghavan-Upfal, Karp-Kleinberg,
% BenOr-Hassidim, Meister-Nietert. Point out
% that the use of exponential weights is
% prefigured in the Burnashev, Karp, Ben-Or
% papers, though all of them are for the
% stochastic setting. Go into detail on the
% results of Meister-Nietert, the technique
% for obtaining the upper bound, the exponential
% gap between upper and lower bounds when
% $\eps$ is constant.
% \item
% Papers Avrim advised Princewill to look at.
% \item
% Anything else? Perhaps streaming algorithms
% for quantile estimation?
% \end{enumerate}

\section{Threshold Query Model}

We start by describing the CDF estimation problem as defined by \citet{meister}. At each time step $t$, the adversary produces a sample $x_t \in [n+1]$, and the algorithm $\mathcal{A}$ generates a query $q_t \in [n]$, each ignorant of the other's choice. Then, the algorithm receives feedback ${\textbf 1}(x_t \leq q_t)$ and produces a CDF estimate $\hat{F}_t$ of $x_1, \ldots, x_t$ while $q_t$ is revealed to the adversary. The adversary is allowed to be adaptive, i.e may select $x_t$ based on the history of the prior $t-1$ time steps. Let $F_t$, defined by $F_t (i) = \frac{1}{t} \sum_{\tau = 1}^t {\textbf 1}(x_\tau \leq i)$ be the empirical CDF of the sequence $x_1, \ldots, x_t$, where $F_t(0) = 0$. They define the estimation error of the algorithm at time $t$ as the Kolmogorov-Smirnov distance between the empirical CDF, $F_t$, and the algorithm's estimate $\hat{F}_t : [n] \rightarrow [0,1]$. In other words, the estimation error is $|| \hat{F}_t - F_t ||_\infty := \sup_{i \in [n]} | \hat{F}_t (i) - F_t (i)|$. 

\vgmargincomment{i added a newline here}
We generalize this formulation in two ways. First, we allow the adversary to pick any monotone function \pcoedit{$v_t : [n] \rightarrow [0,1]$ at each time step instead of a sample $x_t \in [n]$. This generalizes the original setting since $v_t (i) = {\textbf 1}(x_t \leq i)$ is a monotone step function. Thus, picking $x_t \in [n]$ is equivalent to choosing a monotone step function.}
\vgedit{Then, $F_t$ can easily be redefined to be the empirical average of the monotone functions instead, i.e. $F_t(i) = \frac1t \sum_{\tau = 1}^t v_\tau(i)$.}
Second, rather than insisting that the algorithm must make only one query per sample, we \pcodelete{generalize the model to give} \pcoedit{allow} the algorithm a specified number of queries per sample, where this query budget could be anywhere between 1 and $n$.
We now proceed to formalize this Threshold Query Model (TQM). Consider an online estimation environment defined by parameters $n, k$, where the timing of each round $t$ is as follows:
\begin{enumerate}
    \item Adversary chooses a monotone function \vgedit{$v_t : [n] \rightarrow [0,1]$.}
    \item Algorithm chooses a set of (up to) $k$ query points: $q_{1,t} \leq \ldots \leq q_{k,t} \in [n]$; adversary observes these query points.
    \item Algorithm receives feedback: $y_{1,t} \leq \ldots \leq y_{k,t} \in [0,1]$, where $y_{i,t} = \vgedit{v}_t(q_{i,t}).$
\end{enumerate}
We will refer to an environment with this interaction structure as the $k$-TQM,
and we will refer to the parameter $k$ as the {\em query budget}. Since much of our focus is on the case when the query budget equals 1, we will refer to the 1-TQM simply as the TQM.

At the end of the $T$ rounds, the algorithm returns a function $G_T: [n] \rightarrow [0,1]$. We say this algorithm
% is $\varepsilon$-accurate for the $(k, \delta)$-TQM  if it guarantees that $||G_T - \frac{1}{T}\sum F_t||_\infty \leq \varepsilon$ for all $T \geq T_0$ against any adversary or adversarially chosen sequence of $F_t$ and $y_t$.
has {\em accuracy $\eps$} and {\em sample complexity $T$} if it satisfies
the guarantee that for all $\tau \ge T$, the probability that
$\|G_{\tau} - F_{\tau}\| \le \eps$ is at least $\frac34$, against any (potentially adaptive) adversary. For brevity, we will sometimes refer to an algorithm with accuracy $\eps$ and sample complexity $T$ as an $(\eps,T)$-algorithm.
% We we call this a $(\varepsilon, T_0)$-algorithm for the $(k, \delta)$-TQM. \vgcomment{notationally, do we need both T0 and T? could we phrase it as this algo is eps accurate within T steps if for the k, epsilon, infintiy
% norm condition met. i.e. (e, T) algo}
We are interested in the following questions:

\begin{enumerate}
    \item For a fixed accuracy $\eps$ and query budget $k$, what is the minimum sample complexity?
    In other words, what is the smallest $T$ for which there exists an $(\eps, T)$-algorithm for the $k$-TQM?
    \item For a fixed accuracy $\eps$ and sample complexity $T$,
    how large must the query budget be?
    % $\varepsilon, T_0, \delta$, how large must $k$ be?
    In other words, what is the smallest $k$ for which there exists a
    $(\eps,T)$-algorithm for the $k$-TQM?
    % \item For a fixed $\varepsilon, k, T_0$, how small must $\delta$ be? In other words, what is the largest $\delta$ for which there exists a $(\varepsilon, T_0)$-algorithm for the $(k, \delta)$-DQM? Note that $\delta$ cannot be more than $\varepsilon$ since even seeing $y_i$ for all the points at each timestep would not be enough to being $\varepsilon$-accurate.
\end{enumerate}
\rdkedit{For $k=1$, our work shows that the answer to the first question is $O \left( \frac{\log n}{\eps^3} \right)$, which is tight up to a constant factor. 
We also resolve the second question precisely when $T$ is sufficiently large as a function of $n$ and $\eps$, showing that query budget $k = \frac{1}{2 \eps} - 1$ is necessary and sufficient for deterministic algorithms and that query budget $k=1$ is necessary and sufficient for randomized algorithms.}

For simplicity, we shall say a pair $(k, T)$ is \textbf{achievable} if there is an $(\eps,T)$-algorithm for the $k$-TQM.
Algorithms for the $k$-TQM may be deterministic or randomized.
Observe that an $\eps$-accurate deterministic algorithm for the $k$-TQM
must achieve the guarantee that $\|G_{\tau} - F_{\tau}\|_{\infty} \le \eps$
{\em with probability 1} for all $\tau \ge T$ and all adversaries.
% We will say a pair $(k, T_0)$ is \textbf{robustly-achievable} if there an $(\varepsilon, T_0)$-algorithm for the $(k, \varepsilon/4)$-TQM. Clearly, $\delta$ must be less than $\varepsilon$ for this to be achievable. In fact, if $k < n$, $\delta$ must be strictly less than $\delta/2$ \vgmargincomment{epsilon by 2. also, does this condition not hold when k equals n?}.
% For simplicity, we shall say a triple $(k, \delta, T_0)$ (or pair $(k, T_0)$ if $\delta = 0$) is achievable if there a $(\varepsilon, T_0)$-algorithm for the $(k, \delta)$-DQM.\\
% We shall refer to the Threshold Query Model (TQM)%\vgcomment{TQM}
% as a Deterministic Query Model (DQM) when the algorithm's queries $q_{1,t}, \ldots, q_{k, t}$ must be \textit{deterministically} chosen.

\subsection{Deterministic Algorithms for the Threshold Query Model}
The \cite{meister} result can be interpreted as an $\left(\varepsilon, O \left( \frac{n \log n}{\varepsilon^2} \right) \right)$-algorithm for the TQM. In contrast, there does not exist a deterministic algorithm for the TQM. In fact, by using a pigeonhole argument\footnote{The argument is presented in \Cref{pf:deterministic}}
%The adversary always picks a step function any point that is not chosen by the algorithm
% , one can show that for $k < 1/\varepsilon$,
%our impossibility result in %\pcocomment{reference, can also present the argument here},
%
 one can show that deterministic algorithms with accuracy $\eps$ and finite sample complexity must have query budget $k \ge \pcoedit{\frac{1}{2\varepsilon}} - 1$.
% no deterministic algorithm can
% achieve accuracy $\eps$ and finite sample complexity
% in the $k$-TQM
% return an $\varepsilon$-accurate CDF in the $(k, 0)$-TQM.
This shows a qualitative distinction between
% these two models.
deterministic and randomized algorithms for the $k$-TQM, when $k$ is small.

Perhaps the most important reason we study
% the DQM separately from the TQM
deterministic algorithms with query
budget $k>1$ is that it informs our design of a randomized algorithm with query budget 1.
In fact, our algorithms for the TQM (i.e., with query budget 1) work by simulating a deterministic algorithm with a larger query budget.
The algorithm in \cite{meister} can be thought of as simulating a deterministic algorithm with query budget $n$ that simply queries every point at each timestep. This observation suggests a strategy for improving the sample complexity of the algorithm of \citet{meister} by first designing more query-efficient deterministic $k$-TQM algorithms and then simulating them using randomized algorithms.
Our goal is two-fold: making the query budget and sample complexity
simultaneously as small as possible. Naturally, the sample complexity of any simulation algorithm for TQM would depend on these two parameters. However, the smaller the query budget, the greater the sample complexity required for $\eps$-accurate CDF estimation. The trivial deterministic $n$-TQM algorithm guarantees an $\varepsilon$-accurate CDF estimate for every $T \ge 1$, i.e,
its sample complexity is $1$.
In \Cref{sec:elementary}, we show elementary deterministic algorithms,
% $\left( \sqrt{n}, 1/\varepsilon \right)$ and $\left( \log n/\varepsilon, \log n/\varepsilon \right)$ in the DQM.\pcocomment{Bobby, is the phrasing good as is? you can change as you see fit}
one with query budget $O(\sqrt{n}/ \varepsilon)$ and sample complexity
$O(1/\eps)$, the other with query budget $O \left( \frac{\log n}{\eps} \right)$
and sample complexity $O \left( \frac{\log n}{\eps} \right).$
However, neither of these elementary algorithms
attains the lowest possible query budget for deterministic
algorithms. Using the Blackwell's Approachability Theorem, in
\Cref{sec:approach}, we show the existence of a deterministic  algorithm with query budget $O(1/\eps)$, accuracy $\eps$, and finite sample complexity. Then, in \Cref{sec:mult-wt}, we show using the multiplicative weights method that this algorithm can be designed to have sample complexity $O \left( \frac{\log n}{\eps} \right)$.
In \Cref{sec:import-wt}, we adapt this algorithm for the TQM using importance weighting and show that its sample complexity is bounded by
$O \left( \frac{\log n}{\eps^3} \right)$.
In \Cref{sec:low-bd-sketch}, we prove a lower bound on TQM that matches our upper bound up to a constant factor and in \Cref{sec:conclusion}, we consider some generalizations and future work.

\section{Using Approachability to solve TQM}\label{sec:approach}
% \pcocomment{this section is mostly done but needs to be reviewed for errors}

\subsection{Review of Blackwell Approachability}

Blackwell approachability \citep{blackwell} generalizes the problem of playing a repeated two-player zero-sum game to games whose payoffs are vectors instead of scalars. In a Blackwell approachability game, at all times $t$, two players interact in this order: first, Player 1 selects an action $\vgedit{a}_t \in \vgedit{A}$; then, Player 2 selects an action $\vgedit{b}_t \in \vgedit{B}$; finally, Player 1 incurs the vector-valued payoff $\vgedit{h(a_t,b_t)} \in \RR^d$. The sets \vgedit{$A , B$} of player actions are assumed to be compact convex subsets of finite-dimensional vector spaces, and $\vgedit{h}$ is assumed to be a biaffine function on \vgedit{$A \times B$}. Player 1's objective is to guarantee that the average payoff converges to some desired closed convex target set $S \subseteq \RR^d$. Formally, given target set $S \subseteq \RR^d$, Player 1's goal is to pick actions \vgedit{$a_1, a_2, \ldots \in A$} such that no matter the actions \vgedit{$b_1, b_2, \ldots \in B$} played by Player 2,
\begin{equation} \label{eq:approach}
  \dist \left(\frac{1}{T} \sum_{t=1}^T \pcoedit{h}(x_t, y_t), S \right)  \rightarrow 0 \quad \text{as} \quad T \rightarrow \infty
\end{equation}
The action $\vgedit{a}_t$ is allowed to depend on the realized payoff
vectors $\vgedit{h_s(a_s,b_s)}$ for $s=1,2,\ldots,t-1$.
We say the set $S$ is approachable if Player~1
has a strategy that attains the goal~\eqref{eq:approach}
no matter how Player 2 plays. Blackwell's Approachability
Theorem asserts that a convex set $S \subset \reals^d$
is approachable if and only if every closed halfspace
containing $S$ is approachable. This is a
convenient criterion for approachability, because one
can test whether a halfspace is approachable by
computing the value of an associated zero-sum game,
or equivalently by solving a linear program.

\subsection{Approachability Reduction}
A deterministic algorithm for the $k$-TQM
chooses queries $q_{1,t} \le q_{2,t} \le \cdots \le q_{k,t}$ at time $t$, and receives feedback $y_{1,t} \le y_{2,t} \le \cdots \le y_{k,t}$,
where $y_{i,t} = \vgedit{v}_t(q_{i,t})$. For notational convenience, we will interpret $q_{0,t}=0, q_{k+1,t}=n+1, y_{0,t}=0, y_{k+1,t}=1.$ 
% \vgedit{In the approachability reduction, the queries and feedback at time $t$ correspond to the actions taken by Player 1 and 2 respectively.}

\vgmargincomment{i just added a newline here}
For index $i \in [n]$,
% \vgcomment{would prefer adding i in [n] for clarity}
define $\ell_t(i)$ and $u_t(i)$ by
\begin{equation*}
    \ell_t(i) =
    \max \{ y_{j,t}  \mid 0 \le j \le k+1, \, q_{j,t} \le i \}, \qquad
    u_t(i)  =
    \min \{ y_{j,t}  \mid 0 \le j \le k+1, \, q_{j,t} \ge i \}
   .
\end{equation*}
% (The notations $x \vee y$ and $x \wedge y$ denote
% $\max\{x,y\}$ and $\min\{x,y\}$, respectively.) \vgcomment{could we just use max/min then? I think this notation is only used once, in the mult. weights algo description}
% \rdkcomment{No, because I find nested max/min ugly.}
These are the tightest lower and upper bounds on $F_t(i)$
that can be deduced from the values the algorithm queried.
Let $d_t(i) = u_t(i) - \ell_t(i)$.
% \vgedit{In the approachability reduction, the function $d_t: [n] \rightarrow [0, 1]$ corresponds to the vector-valued payoff at time $t$.}
At time $T$, the best lower and upper bounds on
$\frac1T \sum_{t=1}^T \vgedit{v}_t(i)$ that can be deduced
from the values queried are
$\frac1T \sum_{t=1}^T \ell_t(i)$ and
$\frac1T \sum_{t=1}^T u_t(i)$. Hence,
we know that $\frac1T \sum_{t=1}^T \vgedit{v}_t(i)$
belongs to an interval of width $\frac1T \sum_{t=1}^T d_t(i)$.
% When the criterion
% \[
%     \forall i \in [n] \; \;
%     \frac1T \sum_{t=1}^T d_t(i) \le 2 \varepsilon
% \]
% is satisfied,
When this interval width is less than
or equal to $2 \eps$ for every $i \in [n]$,
it is safe to stop and output an estimate
$G_T : [n] \to [0,1]$ defined by setting $G_T(i)$ to be the
midpoint of the interval
$\left[ \frac1T \sum_{t=1}^T \ell_t(i),
\frac1T \sum_{t=1}^T u_t(i) \right]$.

This suggests the following formulation of the deterministic
query model as a game with vector payoff. In each round:
\begin{enumerate}
    \item Adversary chooses monotone non-decreasing
    $\vgedit{v}_t : [n] \to [0,1]$.
    \item Algorithm simultaneously chooses $q_{1,t} \le \cdots \le q_{k,t}$. \vgedit{This corresponds to Player 1's action.}
    \item Feedback $y_{i,t} = \vgedit{v}_t(q_{i,t})$ is revealed for $i=1,2,\ldots,k$. \vgedit{This corresponds to Player 2's action.}
    \item The $n$-dimensional loss vector is $\mathbf{d}_t =
        (d_t(1), \, d_t(2), \, \cdots , \, d_t(n)).$ \vgedit{In our reduction, this corresponds to the vector-valued payoff at time $t$.}
        % \vgcomment{since l and u, (hence d) are previously only defined in terms of the feedback y j,t, should we redefine it in terms of Yt? y j, t = Y t(q j, t), right? } $y_{j, t} = Y_t(q_{j, t})?$
        \vgmargincomment{I can also give the blackwell approachability correspondences in this list}
\end{enumerate}
The questions we seek to understand
are: for which values of $k$ is there an algorithm
that guarantees to approach the set $(-\infty,2 \varepsilon]^n$?
If the set is approachable,  how large must $T$ be to ensure that
the algorithm's $L_{\infty}$ distance from that set is
$O(\varepsilon)$?

\begin{proposition} \label{prop:approachable}
In the vector payoff game corresponding to query budget $k$, the set $(-\infty,2 \varepsilon]^n$ is approachable whenever $k + 1 \ge \frac{1}{2 \varepsilon}$.
\end{proposition}
% \rdkcomment{This seems to violate our earlier claim that
% deterministic $k$-TQM algorithms with finite sample complexity
% do not exist when $k < 1/\eps$. Is the earlier claim off
% by a factor of 2? If so, we should go back and correct it.}
\begin{proof}
    To show that $S = (-\infty,2 \varepsilon]^n$ is an
approachable set, we need to show that every halfspace
containing $S$ is approachable. A halfspace containing
$S$ is a set $H$ of the form
\[
    H = \left\{ \mathbf{x} \left| \sum_{i=1}^n a_i x_i \le b
    \right. \right\}
\]
where $a_1, a_2, \ldots, a_n$ are non-negative,  at least
one of them is strictly positive, and $b \ge
\sum_{i=1}^n a_i (2 \varepsilon).$ Without loss of generality\footnote{%
Otherwise, $H$ is a proper superset of another halfspace
$H'$ that also contains $S$, and to show $H$ is approachable
it suffices to show $H'$ is approachable.},
$b$ is equal to $2 \varepsilon \sum_{i=1}^n a_i$.
%\rdkmargincomment{Moved parenthetical sentence to footnote as recommended by Vaishnavi.} 
Also, without changing the halfspace $H$, we can normalize
$a_1,\ldots,a_n, b$ so that
$\sum_{i=1}^n a_i = 1$ and $b = 2 \varepsilon$.
Assume henceforth that we have adopted such a normalization.

For $j=1,\ldots,k$ let
\begin{equation} \label{eq:qj}
    q_j = \min \left\{ q \, \left| \,
    \sum_{i=1}^q a_i \ge \frac{j}{k+1}
    \right. \right\} .
\end{equation}
We aim to show that for any choice of $\vgedit{v}_t$ by
the adversary, the loss vector $\mathbf{d}_t$ belongs to
$H$ when the algorithm chooses $q_1,\ldots,q_k$ as defined
in Equation~\eqref{eq:qj}.\footnote{%
\vgedit{For brevity, we are using the notation $q_j$ while referring to the query $q_{j, t}$}} For notational convenience,
let $q_0 = 0, \, q_{k+1} = n+1$ and
let $\vgedit{v}_t(0) = 0, \, \vgedit{v}_t(n+1) = 1$. Observe, by the
definition of $q_j$, that
\begin{equation} \label{eq:qj-sum}
    \sum_{i=q_{j}+1}^{q_{j+1}-1} a_i =
    \sum_{i=1}^{q_{j+1}-1} a_i \; - \;
    \sum_{i=1}^{q_{j}} a_i <
    \frac{j+1}{k+1} - \frac{j}{k+1} = \frac{1}{k+1} .
\end{equation}
Also observe that for $i = q_j$
we have $u_t(i) = \ell_t(i) = \vgedit{v}_t(q_j),
\, d_t(i)=0$, while for
$q_j < i < q_{j+1}$ we have
\rdkedit{$u_t(i) = \vgedit{v}_t(q_{j+1})$,
$\ell_t(i) = \vgedit{v}_t(q_j)$,
$d_t(i) = \vgedit{v}_t(q_{j+1}) - \vgedit{v}_t(q_j)$.
Hence,
\begin{align*}
    \sum_{i=1}^n a_i d_t(i)
    & =
    \sum_{j=0}^k \sum_{i=q_j+1}^{q_{j+1}-1} a_i (\vgedit{v}_t(q_{j+1}) - \vgedit{v}_t(q_j)) \\
    & \le
    \sum_{j=0}^k \frac{\vgedit{v}_t(q_{j+1}) - \vgedit{v}_t(q_j)}{k+1}
     = \frac{\vgedit{v}_t(q_{k+1}) - \vgedit{v}_t(q_0)}{k+1}
     = \frac{1}{k+1} .
\end{align*}
}
The right side is less than or equal to $b = 2 \varepsilon$
whenever \rdkedit{$k+1 \ge \frac{1}{2 \eps}.$}
\rdkedit{This confirms that every halfspace containing
$S$ is approachable, hence $S$ is approachable.}
\end{proof}

\section{Multiplicative Weights Algorithm for CDF estimation with parallel queries}\label{sec:mult-wt}

In this section, we transform the proof of approachability
(\Cref{prop:approachable}) into
% an $(\varepsilon,T_0)$-algorithm
% for the $(k,\varepsilon/4)$-DQM whenever
% $k \ge \frac{2}{\varepsilon}$ and
% $T_0 \ge \frac{9 \ln n}{\varepsilon}.$
a deterministic algorithm with query budget $k = \lfloor 1/\eps \rfloor$, accuracy $\eps$,
and sample complexity $\frac{9 \ln  n}{\eps}$.
% In other words, we will prove that the
% pair $(\frac{2}{\varepsilon}, \, \frac{9 \ln n}{\varepsilon})$
% is robustly-achievable.
The key to designing the
algorithm will be to select coefficients
$a_{i,t}$ in each round $t$ using the
multiplicative weights algorithm, and
then respond to these coefficients by
choosing query points $q_{1,t},\ldots,q_{k,t}$
using Equation~\eqref{eq:qj}
as in the proof of approachability.

\begin{algorithm2e}[H]\label{algo:mult-wt}
\caption{Multiplicative weights algorithm for $k$-TQM}
\textbf{Initialize}: \
% $\delta = \varepsilon/4; \eta = 1/3$
\rdkedit{$\eta = 1/3$}\;
$w_{i,0} = \frac1n$ for $i \in [n]$\;
\For{$t=1,2,\ldots,T$}{
    $W = \sum_{i=1}^n w_{i,t-1}$\;
    For $i \in [n]$ let $a_{i,t} = w_{i,t-1}/W$\;
    For $j \in [k]$ let
    $ q_{j,t} = \min \left\{ q \, \left| \,
    \sum_{i=1}^q a_{i,t} \ge \frac{j}{k+1}
    \right. \right\} $\;
    Query points $q_{1,t}, \ldots, q_{k,t}$ and receive
    answers $y_{1,t}, \ldots, y_{k,t}$\;
    Let $q_{0,t} = 0$ and $q_{k+1,t} = n$\;
    \For{$i \in [n]$}{
        \rdkedit{$\ell_t(i) = \max \{ y_{j,t} \mid 0 \le j \le k+1, \, q_{j,t} \le i \}$}\;
        \rdkedit{$u_t(i) =  \min \{ y_{j,t} \mid 0 \le j \le k+1, \, q_{j,t} \ge i \} $}\;
        $d_t(i) = u_t(i) - \ell_t(i)$\;
        $w_{i,t} = w_{i,t-1} \cdot (1+\eta)^{d_t(i)}$\;
    }
}
\textbf{Output}: $G_T[i] = \frac1{2T} \sum_{t=1}^T (\ell_t(i) + u_t(i) )$ for
    all $i \in [n]$.
\end{algorithm2e}

\begin{theorem} \label{thm:mult-wts}
    When $k +1 \ge \frac{1}{\varepsilon}$,
    \rdkedit{Algorithm~\ref{algo:mult-wt} solves the
    $k$-TQM with accuracy $\eps$ and sample
    complexity $\frac{9 \ln n}{\eps}$.}
\end{theorem}
\begin{proof}
    The weights $w_{i,t}$ and coefficients $a_{i,t}$
    in Algorithm~\ref{algo:mult-wt} evolve according to the
    update equations of the standard Hedge algorithm
    with parameter $\eta = 1/3$, and
    according to the analysis of that algorithm in \cite{hedge}
    \rdkmargincomment{Citation needed!},
    we have the inequality
    \begin{equation} \label{eq:hedge}
        \sum_{t=1}^T \sum_{i=1}^n a_i d_t(i) \ge
        (1 - \eta) \max_{i \in [n]} \{ \sum_{t=1}^T d_t(i) \}
        - \frac{\ln n}{\eta} .
    \end{equation}
    From the proof of \Cref{prop:approachable} we know that
    for all $t$, \rdkedit{$\sum_{i=1}^n a_i d_t(i) \le \frac{1}{k+1}$.}
    Substituting this bound into Inequality~\eqref{eq:hedge}
    we obtain
    \rdkedit{%
    \begin{equation} \label{eq:mw1}
        \frac{T}{k+1} \ge
        (1 - \eta) \max_{i \in [n]} \left\{ \sum_{t=1}^T d_t(i) \right\}
        - \frac{\ln n}{\eta}  =
        \frac23 \max_{i \in [n]} \left\{ \sum_{t=1}^T d_t(i) \right\}
        - 3 \ln n \ge
        \frac23 \max_{i \in [n]} \left\{ \sum_{t=1}^T d_t(i) \right\}
        - \frac{\varepsilon}{3} T,
    \end{equation}
    }%
    where the second inequality follows from the fact
    that \rdkedit{$T \ge \frac{9 \ln n}{\varepsilon}$}.
    Recalling that \rdkedit{$\eta = \frac13,
    k+1 \ge \frac{1}{\varepsilon},$} we find that
    \rdkedit{
    \begin{align*}
        \eps T  & \ge
        \frac23 \max_{i \in [n]} \left\{ \sum_{t=1}^T d_t(i) \right\}
        - \frac{\varepsilon}{3} T  \\
        \frac43 \eps T & \ge
        \frac23 \max_{i \in [n]} \left\{ \sum_{t=1}^T d_t(i) \right\} \\
        2 \varepsilon & \ge
        \max_{i \in [n]} \left\{ \tfrac 1T \sum_{t=1}^T d_t(i) \right\} .
    \end{align*}
    }%
    The right side of the last inequality is equal to the
    width of the interval
    $\left[  \frac1T \sum_{t=1}^T \ell_t(i), \;
    \frac1T \sum_{t=1}^T u_t(i) \right]$.
    That interval is guaranteed to contain
    $\frac1T \sum_{t=1}^T \vgedit{v}_t(i)$, and its
    midpoint is $G_T(i)$, so we are assured
    that
    \rdkedit{$|G_T(i) - \frac1T \sum_{t=1}^T 
    \vgedit{v}_t(i)|
    \le \frac12 \left( 2 \eps \right) = \eps$,}
    as desired.
\end{proof}

\section{Randomized algorithm using importance weighting}\label{sec:import-wt}

In \Cref{sec:mult-wt}, we presented a deterministic algorithm
% that returns an $\varepsilon$-accurate estimate of the CDF after $\frac{9\ln n}{\varepsilon}$ time-steps using only $\frac{2}{\varepsilon}$ queries per time-step.
with accuracy $\eps$, query budget $\lfloor \frac{1}{\eps} \rfloor$,
and sample complexity $\frac{9 \ln n}{\eps}$.
 Earlier we noted that no deterministic algorithm can obtain
 % an $\varepsilon$-accurate CDF estimate using less than $\frac{1}{\varepsilon}$ samples.
 accuracy $\eps$ with a query budget less than $\frac{1}{2 \varepsilon} - 1$.
 \rdkmargincomment{Off by a factor of 2?}
 In this section, we turn our attention to \textit{randomized} algorithms
 %that are only allowed to make one query per timestep.
with query budget 1.

One natural approach would be to simulate the deterministic algorithm from the previous section; that is, run one step of Algorithm $\ref{algo:mult-wt}$ to obtain a set of $k$ query points, and over the course of the next
$O\left(\frac{k \log n}{\varepsilon^2} \right)$
time-steps, randomly sample one of the $k$ points to query to get an $\varepsilon$-accurate estimate of the CDF at each of the $k$ query points.
This approach can be carried out successfully, although we omit the
details from this paper. However, the resulting sample complexity bound
exceeds the optimal bound by a factor of $\Omega \left(
\frac{\ln \ln n}{\varepsilon} \right).$
One of the main reasons for this is that the algorithm commits to sampling from a \textit{fixed} set of $k$ points for $O\left(\frac{k \log n}{\varepsilon^2} \right)$ time steps even though the algorithm's CDF estimate is changing and Algorithm~\ref{algo:mult-wt} might suggest a different set of $k$ query points.
%\pcocomment{Consider splitting up last sentence}

To circumvent this issue of committing to a fixed set of query points, we use an approach from the bandit literature known as importance weighting.
% importance weighted estimates of the timestep cdf at the k query points .
% Simply put, we replace the feedback $y_{1,t} \leq \ldots \leq y_{k,t}$ with % importance weighted estimates $\hat{y}_{1,t} \leq \ldots \leq \hat{y}_{k,t}$.
We can't query all the points $q_{1,t}, \ldots, q_{k,t}$ to receive feedback $y_{1,t}, \ldots, y_{k,t}$, so we instead choose one point uniformly at random $q_{m,t}$, to receive feedback $y_{m,t}$. We set the values $\hat{y}_{j,t}$ to $k \cdot y_{m,t}$ if $j = m$ and $0$ otherwise. Then we proceed with the rest of the algorithm with values $\hat{y}_{1,t}, \ldots, \hat{y}_{k,t}$ instead of $y_{1,t}, \ldots, y_{k,t}$.

%\rdkcomment{Modified the following pseudocode to omit $\delta$'s.}

\begin{algorithm2e}[H]\label{algo:import-wt}
\caption{Randomized MW algorithm using importance weighting}
\textbf{Initialize}: \
% \rdkdelete{$\delta = \varepsilon/4$;}
$k = \frac{2}{\varepsilon}, \eta = \varepsilon^2 /16 $\;
$w_{i,0} = \frac1n$ for $i \in [n]$\;
\For{$t=1,2,\ldots,T$}{
    $W = \sum_{i=1}^n w_{i,t-1}$\;
    For $i \in [n]$ let $a_{i,t} = w_{i,t-1}/W$\;
    For $j \in [k]$ let
    $ q_{j,t} = \min \left\{ q \, \left| \,
    \sum_{i=1}^q a_{i,t} \ge \frac{j}{k+1}
    \right. \right\} $\;
    Let $q_{0,t} = 0$ and $q_{k+1,t} = n$\;
    For $j \in [k]$ let $\hat{y}_{j,t} = 0$\;
    Sample $m$ uniformly from $[k]$\;
    Query $q_{m,t}$ and receive $y_{m,t}$\;
    Set $\hat{y}_{m,t} = k \cdot y_{m,t}$\;
    \For{$i \in [n]$}{
        $\hat{\ell}_t(i) = \hat{y}_{r,t} \ \text{where} \ r = \max \{ j \mid 0 \le j \le k+1, \, q_{j,t} \le i \} $\;
        $\hat{u}_t(i) = \hat{y}_{r,t} \ \text{where} \ r = \min \{ j \mid 0 \le j \le k+1, \, q_{j,t} \ge i \} $\;
        $\hat{d}_t(i) = \hat{u}_t(i) - \hat{\ell}_t(i)$\;
        $w_{i,t} = w_{i,t-1} \cdot (1+\eta)^{\hat{d}_t(i)}$\;
    }
}
\textbf{Output}: $\hat{G}_T[i] = \frac1{2T} \sum_{t=1}^T (\hat{\ell}_t(i) + \hat{u}_t(i) )$ for all $i \in [n]$.
\end{algorithm2e}

\rdkcomment{Should the output be denoted by $G_T$ rather
than $\hat{G}_T$?}

\begin{theorem}\label{thm:import-wts}
    Algorithm~\ref{algo:import-wt}
    solves the TQM with accuracy $\eps$ and
    sample complexity $\frac{64 \log n}{\eps^3}$.
\end{theorem}
The proof of the theorem is presented in
\Cref{sec:import-wts-proof}.

\section{Lower Bound}
\label{sec:low-bd-sketch}

In this section we sketch a proof that the sample
complexity of \Cref{algo:import-wt} is
information-theoretically optimal, up to
a constant factor. The full proof appears
in \Cref{sec:low-bd-proof}.

\begin{theorem}  \label{thm:low-bd}
  For any $n > 1, \eps > 0$, every algorithm that solves
  the CDF estimation problem with accuracy $\eps$, using
  one threshold query per sample, requires at least
  $T_0 = \Omega(\min\{n,\frac{1}{\eps} \cdot \log(n) / \eps^2)$
  samples.
\end{theorem}

When $n \le \frac{1}{\eps}$ this is Theorem 6
of \citet{meister}, so for the remainder of this
section we discuss the proof when $n > \frac{1}{\eps}.$
Assume for convenience\footnote{These
assumptions are without loss of generality.
First increase $\eps$ by a factor of at most 6,
to ensure that $\frac{1}{6\eps}$ is an integer less than or
equal to $n$, then decrease $n$ by a factor of
at most 2 to ensure that $n$ is divisible by $\frac{1}{6 \eps}$.
These changes to $\eps$ and $n$ only affect the implicit constant
in the big-$\Omega$ expression for the lower bound.}
that the numbers
% \[
%   k = \tfrac{1}{6 \eps}
%     \quad \mbox{and} \quad
%   m = \tfrac{n}{k} = 6 \eps n
% \]
$k = \frac{1}{6 \eps}$ and $m = n/k = 6 \eps n$
are positive integers. We will then define a
family of probability distributions parameterized
by $\theta \in [m]^k$. Their cumulative distribution
functions, $\{ F_{\theta} \mid \theta \in [m]^k\}$,
are designed to have three properties.
\begin{enumerate}
  \item For any function $\hat{F}$ there is at most
    one $\theta \in [m]^k$ such that
    $\| \hat{F} - F_{\theta} \|_{\infty} < \frac{3 \eps}{2}.$
    Hence, given a $\frac{3 \eps}{2}$-accurate estimate of $F_{\theta}$
    we can deduce the value of $\theta$.
    (\Cref{lem:ftheta-decode})
  \item Informally, any sequence of $1/\eps^{2}$
    threshold queries reveals at most $O(1)$ bits
    of information about $\theta$. In the proof,
    this statement is formalized information-theoretically
    in terms of the expected KL-divergence between the
    learner's prior and posterior distributions over
    $\theta$.
    (\Cref{lem:klbayes-ub})
  \item Starting from a uniform prior over
    $\theta \in [m]^k$, the posterior distribution
    after any sequence of threshold queries is a
    product distribution. In other words, writing
    $\theta$ as a $k$-tuple $(\theta_1,\ldots,\theta_k),$
    the $k$ coordinates of the tuple are mutually
    independent under the posterior distribution.
    (\Cref{lem:posterior-is-product})
\end{enumerate}
Now suppose the adversary generates a sequence
$x_1,\ldots,x_T$ by sampling $\theta \in [m]^k$
uniformly at random and then drawing $T$
independent samples from the distribution
with CDF $F_{\theta}.$
Using the Dvoretzky-Kiefer-Wolfowitz Inequality,
we will argue that with probability at least $\frac{7}{8}$,
the empirical CDF of the samples differs from
$F_{\theta}$ by less than $\frac{\eps}{2}$ in
$L_{\infty}$ norm. Hence, an $\eps$-accurate
estimate of the empirical distribution is a
$\frac{3 \eps}{2}$-accurate estimate of $F_{\theta}$,
and consequently it uniquely determines the value
of $\theta$. This implies that an algorithm which
succeeds, with probability at least $\frac34$,
in outputting an $\eps$-accurate estimate of
the empirical distribution of the samples,
must also succeed with probability at least
$\frac58$ in learning the exact value of
$\theta$, a random variable of entropy
$k \log(m)$. Since it takes $\Omega(1/\eps^{2})$
queries to learn a single bit of information
about $\theta$, it takes $\Omega(k \log(m) / \eps^2)$
queries to learn the exact value of $\theta$ with
constant probability. Recalling the definitions
of $k$ and $m$, we see that this bound is
$\Omega(\log(\eps n) / \eps^3).$ \Cref{thm:low-bd}
asserts the stronger lower bound $\Omega(\log(n) / \eps^3)$,
which is asymptotically greater when $1/\eps = n^{1-o(1)}.$
To strengthen the lower bound in this case, we use the third
property of the construction --- that the posterior
distribution over $\theta$ is a product distribution ---
to prove a stronger lower bound on the expected KL divergence
between the prior and posterior distributions at the time
when the algorithm outputs its estimate.

\section{Discussion and Open Problems}\label{sec:conclusion}
\pcoedit{
In addressing the online CDF estimation problem, we took a detour to a more general setting - the Threshold Query Model. Although we completely characterize the sample complexity of online CDF estimation using threshold queries, this only resolves the sample complexity question for the $k$-TQM for $k = 1$. One direction for future work is to characterize the optimal sample complexity for every value of the query budget $k$. Another direction is to extend the Threshold Query Model to other distance metrics besides the Kolmogorov-Smirov distance. It is important to note that the presented algorithms are fully adaptive. In some practical settings, however, there might issues of latency and delays. This raises a question of whether the query budget and sample complexity is higher for non-adaptive algorithms, and if so, by how much? Some special cases of this are addressed in \Cref{sec:elementary}.

A natural extension to consider is the continuous-support setting, where the queries and samples can be any real value in the interval $[0,1]$. Without any additional assumptions, CDF estimation in this setting becomes intractable. This is because there are infinitely many values and the algorithm cannot cover all of them with a finite number of queries. However, \cite{meister} point out that if we specify some resolution $r$ of interest, then by setting $n = O(1/r)$, this reduces to the discrete-support case. We wonder what assumptions on the adversary's probability density function would make the CDF estimation problem tractible and if our techniques would be applicable.
% Another assumption that could be made is to require that the adversary's probability density function for each timestep is bounded at every point in the interval. This setting turns out to be equivalent to the discrete-support. It would be useful to see if CDF estimation can be done in the continuous-support setting with weaker assumptions.
%We provide a sketch of the argument here.
\vgedit{Another direction of interest is to consider higher-dimensional generalizations that estimate multivariate distributions - the samples are now vector-valued, and the algorithm queries the data using linear threshold functions (i.e., halfspaces).}

% \vgedit{Further analysis of the $k$-TQM is interesting in its own right, and future directions for analysis can involve characterizing the optimal sample complexity for every value of the query budget $k$ (not just $k = 1$), and using metrics other than the Kolmogorov-Smirnov distance. Behaviour with a more constrained adversary (for example, fewer rounds of adaptivity) can also be analysed. We explore some special cases in \Cref{sec:elementary}}.

In \Cref{sec:intro}, we pointed out some related works in \vgedit{relevant application areas like} online dynamic pricing and auctions. We wonder if our techniques can be extended to these settings as well.}

\bibliographystyle{apalike}
\bibliography{bibliography}

\begin{thebibliography}{}

\bibitem[Acharya et~al., 2021]{acharya-multiple-samples}
Acharya, J., Canonne, C., Liu, Y., Sun, Z., and Tyagi, H. (2021).
\newblock Distributed estimation with multiple samples per user: Sharp rates
  and phase transition.
\newblock In Ranzato, M., Beygelzimer, A., Dauphin, Y., Liang, P., and Vaughan,
  J.~W., editors, {\em Advances in Neural Information Processing Systems},
  volume~34, pages 18920--18931. Curran Associates, Inc.

\bibitem[Acharya et~al., 2020]{acharya-information-constraints}
Acharya, J., Canonne, C.~L., Sun, Z., and Tyagi, H. (2020).
\newblock Unified lower bounds for interactive high-dimensional estimation
  under information constraints.
\newblock {\em CoRR}, abs/2010.06562.

\bibitem[Ali and Ronaldson, 2012]{ali2012}
Ali, S. and Ronaldson, S. (2012).
\newblock Ordinal preference elicitation methods in health economics and health
  services research: using discrete choice experiments and ranking methods.
\newblock {\em British medical bulletin}, 103(1):21--44.

\bibitem[Arora et~al., 2012]{hedge}
Arora, S., Hazan, E., and Kale, S. (2012).
\newblock The multiplicative weights update method: a meta-algorithm and
  applications.
\newblock {\em Theory of Computing}, 8(6):121--164.

\bibitem[Auer et~al., 2002]{auer}
Auer, P., Cesa-Bianchi, N., Freund, Y., and Schapire, R.~E. (2002).
\newblock The nonstochastic multiarmed bandit problem.
\newblock {\em SIAM Journal on Computing}, 32(1):48--77.

\bibitem[Barlow et~al., 1972]{isotonic}
Barlow, R.~E., Bartholomew, D.~J., Bremner, J.~M., and Brunk, H.~D. (1972).
\newblock {\em Statistical inference under order restrictions: The theory and
  application of isotonic regression}.
\newblock Wiley.

\bibitem[Barnes et~al., 2020]{barnes}
Barnes, L.~P., Han, Y., and Ozgur, A. (2020).
\newblock Lower bounds for learning distributions under communication
  constraints via fisher information.
\newblock {\em Journal of Machine Learning Research}, 21(236):1--30.

\bibitem[Ben-Or and Hassidim, 2008]{benor}
Ben-Or, M. and Hassidim, A. (2008).
\newblock The bayesian learner is optimal for noisy binary search (and pretty
  good for quantum as well).
\newblock In {\em 2008 49th Annual IEEE Symposium on Foundations of Computer
  Science}, pages 221--230. IEEE.

\bibitem[Blackwell, 1956]{blackwell}
Blackwell, D. (1956).
\newblock {An analog of the minimax theorem for vector payoffs.}
\newblock {\em Pacific Journal of Mathematics}, 6(1):1 -- 8.

\bibitem[Blum et~al., 2015]{blum}
Blum, A., Mansour, Y., and Morgenstern, J. (2015).
\newblock Learning valuation distributions from partial observations.
\newblock In {\em Proceedings of the Twenty-Ninth AAAI Conference on Artificial
  Intelligence}, AAAI'15, page 798–804. AAAI Press.

\bibitem[Burnashev and Zigangirov, 1974]{burnashev}
Burnashev, M.~V. and Zigangirov, K.~S. (1974).
\newblock An interval estimation problem for controlled observations.
\newblock {\em Problems of Information Transmission}, 10:223--231.
\newblock (published in Russian).

\bibitem[Chapelle et~al., 2012]{chapelle}
Chapelle, O., Joachims, T., Radlinski, F., and Yue, Y. (2012).
\newblock Large-scale validation and analysis of interleaved search evaluation.
\newblock {\em ACM Transactions on Information Systems (TOIS)}, 30(1):1--41.

\bibitem[Cormode et~al., 2004]{cormode2004holistic}
Cormode, G., Johnson, T., Korn, F., Muthukrishnan, S., Spatscheck, O., and
  Srivastava, D. (2004).
\newblock Holistic udafs at streaming speeds.
\newblock In {\em Proceedings of the 2004 ACM SIGMOD international conference
  on Management of data}, pages 35--46.

\bibitem[Dvoretzky et~al., 1956]{dkw}
Dvoretzky, A., Kiefer, J., and Wolfowitz, J. (1956).
\newblock {Asymptotic Minimax Character of the Sample Distribution Function and
  of the Classical Multinomial Estimator}.
\newblock {\em The Annals of Mathematical Statistics}, 27(3):642 -- 669.

\bibitem[Feige et~al., 1994]{feige}
Feige, U., Raghavan, P., Peleg, D., and Upfal, E. (1994).
\newblock Computing with noisy information.
\newblock {\em SIAM Journal on Computing}, 23(5):1001--1018.

\bibitem[Freedman, 1975]{freedman}
Freedman, D.~A. (1975).
\newblock {On Tail Probabilities for Martingales}.
\newblock {\em The Annals of Probability}, 3(1):100 -- 118.

\bibitem[Greenwald and Khanna, 2001]{greenwald2001space}
Greenwald, M. and Khanna, S. (2001).
\newblock Space-efficient online computation of quantile summaries.
\newblock {\em ACM SIGMOD Record}, 30(2):58--66.

\bibitem[Han et~al., 2018]{han}
Han, Y., {\"{O}}zg{\"{u}}r, A., and Weissman, T. (2018).
\newblock Geometric lower bounds for distributed parameter estimation under
  communication constraints.
\newblock In {\em Proceedings of the 31st Conference on Learning Theory, {COLT}
  2018}, volume~75 of {\em Proceedings of Machine Learning Research}, pages
  3163--3188. {PMLR}.
\newblock The arXiv (v3) version from 2020 corrects some issues and includes
  more results.

\bibitem[Karnin et~al., 2016]{karnin2016optimal}
Karnin, Z., Lang, K., and Liberty, E. (2016).
\newblock Optimal quantile approximation in streams.
\newblock In {\em 2016 ieee 57th annual symposium on foundations of computer
  science (focs)}, pages 71--78. IEEE.

\bibitem[Karp and Kleinberg, 2007]{karp}
Karp, R.~M. and Kleinberg, R. (2007).
\newblock Noisy binary search and its applications.
\newblock In {\em Proc. 18th ACM-SIAM Symposium on Discrete Algorithms (SODA)},
  pages 881--890.

\bibitem[Kleinberg and Leighton, 2003]{kleinberg}
Kleinberg, R. and Leighton, T. (2003).
\newblock The value of knowing a demand curve: bounds on regret for online
  posted-price auctions.
\newblock In {\em 44th Annual IEEE Symposium on Foundations of Computer
  Science, 2003. Proceedings.}, pages 594--605.

\bibitem[Larichev et~al., 1995]{larichev}
Larichev, O., Olson, D., Moshkovich, H., and Mechitov, A. (1995).
\newblock Numerical vs cardinal measurements in multiattribute decision making:
  How exact is enough?
\newblock {\em Organizational Behavior and Human Decision Processes},
  64(1):9--21.

\bibitem[Leme et~al., 2021a]{leme}
Leme, R.~P., Sivan, B., Teng, Y., and Worah, P. (2021a).
\newblock Learning to price against a moving target.
\newblock In Meila, M. and Zhang, T., editors, {\em Proceedings of the 38th
  International Conference on Machine Learning}, volume 139 of {\em Proceedings
  of Machine Learning Research}, pages 6223--6232. PMLR.

\bibitem[Leme et~al., 2021b]{reviewer1request}
Leme, R.~P., Sivan, B., Teng, Y., and Worah, P. (2021b).
\newblock Pricing query complexity of revenue maximization.
\newblock {\em CoRR}, abs/2111.03158.

\bibitem[Manku et~al., 1998]{manku1998approximate}
Manku, G.~S., Rajagopalan, S., and Lindsay, B.~G. (1998).
\newblock Approximate medians and other quantiles in one pass and with limited
  memory.
\newblock {\em ACM SIGMOD Record}, 27(2):426--435.

\bibitem[Meister and Nietert, 2021]{meister}
Meister, M. and Nietert, S. (2021).
\newblock Learning with comparison feedback: Online estimation of sample
  statistics.
\newblock In {\em Algorithmic Learning Theory}, pages 983--1001. PMLR.

\bibitem[Moshkovich et~al., 2002]{moshkovich}
Moshkovich, H.~M., Mechitov, A.~I., and Olson, D.~L. (2002).
\newblock Ordinal judgments in multiattribute decision analysis.
\newblock {\em European Journal of Operational Research}, 137(3):625--641.

\bibitem[Munro and Paterson, 1980]{munro1980selection}
Munro, J.~I. and Paterson, M.~S. (1980).
\newblock Selection and sorting with limited storage.
\newblock {\em Theoretical computer science}, 12(3):315--323.

\bibitem[Perchet, 2015]{perchet}
Perchet, V. (2015).
\newblock Exponential weight approachability, applications to calibration and
  regret minimization.
\newblock {\em Dynamic Games and Applications}, 5:136--153.

\bibitem[Shrivastava et~al., 2004]{shrivastava2004medians}
Shrivastava, N., Buragohain, C., Agrawal, D., and Suri, S. (2004).
\newblock Medians and beyond: new aggregation techniques for sensor networks.
\newblock In {\em Proceedings of the 2nd international conference on Embedded
  networked sensor systems}, pages 239--249.

\bibitem[Weed et~al., 2016]{weed16-auctions}
Weed, J., Perchet, V., and Rigollet, P. (2016).
\newblock Online learning in repeated auctions.
\newblock In Feldman, V., Rakhlin, A., and Shamir, O., editors, {\em 29th
  Annual Conference on Learning Theory}, volume~49 of {\em Proceedings of
  Machine Learning Research}, pages 1562--1583, Columbia University, New York,
  New York, USA. PMLR.

\end{thebibliography}

\appendix
\section{Proof of Lower Bound}
\label{sec:low-bd-proof}

Throughout this section we assume $n > \frac{1}{\eps}$,
since the case $n \le \frac{1}{\eps}$ of \Cref{thm:low-bd}
was already proven by \citet{meister}. Additionally, as
explained in \Cref{sec:low-bd-sketch}, we assume without
loss of generality that $n = km,$ where $k = \frac{1}{6 \eps}.$

We begin in \Cref{sec:lb-family}
by describing the construction
of a family of distributions over $[n]$
parameterized by $\theta \in [m]^k$.
Then, in \Cref{sec:kl} we review some basic
facts about KL divergence, the main
information theoretic tool in the proof.
Finally, \Cref{sec:lb-finish} completes the
proof.

\subsection{Family of distributions} \label{sec:lb-family}

Since $n=km$, each element of $[n]$ can be uniquely expressed
in the form $(a-1) m + b$ where $a \in [k]$ and $b \in [m].$
For $\theta = (\theta_1,\theta_2,\ldots,\theta_{k}) \in [m]^k$
let $I_{\theta}$ denote the set
\[
  I_{\theta} = \{ (a-1) m + \theta_a \mid a \in [k] \}
\]
and let $D_{\theta}$ denote the probability distribution
on $[n]$ representing the output of the following
sampling rule.
\begin{enumerate}
  \item With probability $\frac12$ output a uniformly random
    element of $I_{\theta}$.
  \item With probability $\frac14$ output 1.
  \item With probability $\frac14$ output $n$.
\end{enumerate}
The cumulative distribution function of $D_{\theta}$ is
the function $F_{\theta}$ whose value
at $i = (a-1) m + b$, when $a \in [k], \, b \in [m]$,
is given by the formula
\begin{equation} \label{eq:ftheta}
  F_{\theta}((a-1)m + b) = \begin{cases}
    \frac14 + 3 (a-1) \eps & \mbox{if } b < \theta_a \\
    \frac14 + 3 a \eps & \mbox{if } b \ge \theta_a, \, (a-1)m + b < n \\
    1 & \mbox{if } (a-1)m + b =  n .
  \end{cases}
\end{equation}
\begin{lemma} \label{lem:ftheta-close}
  If $\theta, \theta'$ are any two distinct elements of $[m]^k$
  then $\| F_{\theta} - F_{\theta'} \|_{\infty} = 3 \eps$.
\end{lemma}
\begin{proof}
  From the definition of $F_{\theta}$ and $F_{\theta'}$ it
  is apparent that $F_{\theta}(n) = F_{\theta'}(n) = 1$ and
  that for $i = (a-1)m + b < n$, $F_{\theta}(i)$ and
  $F_{\theta'}(i)$ both belong to the set
  $\{\frac14 + 3 (a-1) \eps, \, \frac14 + 3 a \eps\},$
  so $|F_{\theta}(i) - F_{\theta'}(i)|$ cannot
  exceed $3 \eps$. Thus $\|F_{\theta} - F_{\theta'}\|_{\infty} \le
  3 \eps$.

  Since we are assuming $\theta \neq \theta'$,
  there exists some $a \in [k]$ such that $\theta_a \neq \theta'_a$.
  Assume without loss of generality that $\theta_a < \theta'_a$.
  Then, $i = (a-1) m + \theta_a,$ we have
  $F_{\theta}(i) - F_{\theta'}(i) = 3 \eps$.
  Thus, $\|F_{\theta} - F_{\theta'}\|_{\infty} \ge 3 \eps.$
\end{proof}

\begin{lemma} \label{lem:ftheta-decode}
  For any function $\hat{F} : [n] \to [0,1]$ there is at
  most one $\theta \in [m]^k$ satisfying
  $\| \hat{F} - F_{\theta} \|_{\infty} < \frac{3 \eps}{2} .$
\end{lemma}
\begin{proof}
  The lemma follows immediately from \Cref{lem:ftheta-close}
  and fact that the $L_{\infty}$ norm satisfies the triangle inequality.
\end{proof}

In the proof to follow, we will be considering executing a
fixed but arbitrary CDF estimation algorithm on an input
sequence $x_1,x_2,\ldots,x_T$ generated as follows: first
sample $\theta \in [m]^k$ uniformly at random, then let
$x_1,x_2,\ldots,x_T$ be $T$ independent samples
from $D_{\theta}$. Let $(q_1,y_1),(q_2,y_2),\ldots,(q_T,y_T)$ be
the random $[n] \times \{0,1\}$-valued sequence representing
the algorithm's queries and the responses to those queries.
For $0 \le t \le T$ let $p_t(\theta)$ denote the
posterior distribution of $\theta$ given $(q_1,y_1),\ldots,(q_t,y_t)$.
(When $t=0$ this is simply the prior distribution of
$\theta$, i.e.~the uniform distribution on $[m]^k$.)

\begin{lemma} \label{lem:posterior-is-product}
  For $0 \le t \le T$, for all sequences of queries
  and responses $(q_1,y_1), \ldots, (q_t,y_t),$
  the posterior distribution $p_t$ is a product distribution.
  % In other words, there exist functions
  % $f_a : [m] \times ([n] \times \{0,1\})^t \to [0,1]$ such that
  % for every sequence of queries and responses, $(q_1,y_1),\ldots,(q_t,y_t)$
  % and for every $\theta = (\theta_1,\ldots,\theta_k) \in [m]^k$,
  % \[
  %   p_t(\theta \mid (q_1, y_1), (q_2, y_2), \ldots, (q_t, y_t)) =
  %   \prod_{a=1}^k f_a(\theta_a, (q_1, y_1), (q_2, y_2), \ldots, (q_t, y_t)) .
  % \]
  In other words, if $\theta = (\theta_1,\ldots,\theta_k) \in [m]^k$
  is distributed according to $p_t$ then the random variables
  $\theta_1,\ldots,\theta_k$ are mutually independent.
\end{lemma}
\begin{proof}
  The proof is by induction in $t$. In the base case,
  $p_0$ is the uniform distribution on $[m]^k$, which
  is a product distribution. For the induction step,
  write $q_t \in [n]$ as $q_t = (a_t - 1) m + b_t$
  where $a_t \in [k], b_t \in [m]$. The
  distribution of $y_t$ given $q_t$ depends only
  on the parameter $\theta_{a_t}.$ Therefore,
  if $p_{t-1}$ is a product distribution,
  a Bayesian update
  % of the distribution of
  % $\theta = (\theta_1,\ldots,\theta_k)$ given
  conditioning on
  $(q_t,y_t)$ will alter the marginal distribution
  of $\theta_{a_t}$ while leaving it independent
  of $\theta_j$ for all $j \neq a_t.$
\end{proof}

\subsection{Review of KL divergence} \label{sec:kl}

For two probability distributions $p,q$ on a finite set $\Omega$,
their Kullback-Leibler divergence, henceforth called KL divergence,
is defined as
\begin{equation} \label{eq:kl-definition}
  \kldiv{p}{q} = \sum_{\omega \in \Omega} p(\omega) \ln \left( \frac{p(\omega)}{q(\omega)} \right) .
\end{equation}
When $p(\omega)=0$ the summand on the right side
of~\eqref{eq:kl-definition} is interpreted as zero.

In this section and the following one, if $p$ is a
distribution over pairs $(X,Y)$ then $p(X), \, p(Y)$
denote the marginal distribution of $X$ and $Y$, respectively,
and $p(X|Y), \, p(Y|X)$ denote the conditional distribution
of $X$ given $Y$ and the conditional distribution of $Y$
given $X$, respectively.

\begin{lemma} \label{lem:kl-nonneg}
  If $p,q$ are any two probability distributions on the same set,
  then $\kldiv{p}{q} \ge 0.$
\end{lemma}
\begin{proof}
  The function $\phi(x) = -\ln(x)$ is convex, so by
  Jensen's inequality,
  \begin{align*}
    \sum_{\omega \in \Omega} p(\omega) \ln \left( \frac{p(\omega)}{q(\omega)}
    \right)
    & =
    \sum_{\omega \in \Omega} p(\omega) \phi \left( \frac{q(\omega)}{p(\omega)}
    \right) \\
    & \ge
    \phi \left( \sum_{\omega \in \Omega} p(\omega) \cdot \frac{q(\omega)}{p(\omega)}
    \right) \\
    & =
    \phi \left( \sum_{\omega \in \Omega} q(\omega) \right)
    = \phi(1) = 0 .
  \end{align*}
\end{proof}

\subsubsection{The chain rule and its corollaries}
\label{sec:chainrule}

\begin{lemma}[Chain rule for KL divergence] \label{lem:kl-chainrule}
  If $p,q$ are probability distributions on pairs $(X,Y)$ then
  \begin{equation} \label{eq:kl-chainrule}
    \kldiv{p}{q} = \kldiv{p(Y)}{q(Y)} + \expect_{Y \sim p} \kldiv{p(X|Y)}{q(X|Y)} .
  \end{equation}
\end{lemma}
\begin{proof}
  For any $y$, Bayes' Law implies
  \begin{align*}
    \ln p(X=x | Y=y) & = \ln p(X=x, \, Y=y) - \ln p(Y=y) \\
    \ln q(X=x | Y=y) & = \ln q(X=x, \, Y=y) - \ln q(Y=y) \\
    \ln \left( \frac{p(X=x | Y=y)}{q(X=x | Y=y)} \right) & =
    \ln \left( \frac{p(X=x,\,  Y=y)}{q(X=x, \, Y=y)} \right) -
    \ln \left( \frac{p(Y=y)}{q(Y=y)}) \right)
  \end{align*}
  hence
  \begin{align*}
    \expect_{Y \sim p} \kldiv{p(X|Y)}{q(X|Y)} & =
    \sum_{y} p(Y=y) \sum_{x} p(X=x | Y=y) \ln \left(
      \frac{p(X=x | Y=y)}{q(X=x | Y=y)} \right) \\
    & =
    \sum_{x,y} p(X=x, Y=y) \ln \left(
      \frac{p(X=x | Y=y)}{q(X=x | Y=y)} \right) \\
    & =
    \sum_{x,y} p(X=x, Y=y) \ln \left( \frac{p(X=x,\,  Y=y)}{q(X=x, \, Y=y)} \right)
    \; - \;
    \sum_{x,y} p(X=x, Y=y) \ln \left( \frac{p(Y=y)}{q(Y=y)}) \right) \\
    & =
    \sum_{x,y} p(X=x, Y=y) \ln \left( \frac{p(X=x,\,  Y=y)}{q(X=x, \, Y=y)} \right)
    \; - \;
    \sum_y p(Y=y) \ln \left( \frac{p(X=x,\,  Y=y)}{q(X=x, \, Y=y)} \right)  \\
    & = \kldiv{p}{q} - \kldiv{p(Y)}{q(Y)} .
  \end{align*}
\end{proof}
The chain rule has several corollaries which will be of use to us.
\begin{lemma} \label{lem:iter-chainrule}
  If $p,q$ are probability distributions on $t$-tuples
  $X_1,\ldots,X_t$, then
  \begin{equation} \label{eq:iter-chainrule}
    \kldiv{p}{q} = \sum_{s=1}^t \expect_{X_1,\ldots,X_{s-1} \sim p}
      \kldiv{p(X_s | X_1,\ldots,X_{s-1})}{q(X_s | X_1,\ldots,X_{s-1})} .
  \end{equation}
\end{lemma}
\begin{proof}
  The lemma follows by induction on $t$, which the base case $t=1$
  being trivial and the induction step being a direct application
  of \Cref{lem:kl-chainrule}.
\end{proof}
\begin{lemma} \label{lem:kl-additive}
  If $p,q$ are probability distributions on $t$-tuples
  $X_1,\ldots,X_t$, and both $p$ and $q$ are product distributions, then
  \begin{equation} \label{eq:kl-additive}
    \kldiv{p}{q} = \sum_{s=1}^t \kldiv{p(X_s)}{q(X_s)} .
  \end{equation}
\end{lemma}
\begin{proof}
  Since $p$ is a product distribution,
  $p(X_s | X_1,\ldots,X_{s-1}) = p(X_s)$, and
  similarly for $q$.
  The lemma now follows directly from \Cref{lem:iter-chainrule}.
\end{proof}
\begin{lemma} \label{lem:kl-bayesian}
  If $p,q$ are probability distributions on pairs $(X,Y)$
  that have the same marginals --- i.e.,
  $p(X) = q(X)$ and $p(Y) = q(Y)$ --- then
  \begin{equation} \label{eq:kl-bayesian}
    \expect_{X \sim p} \kldiv{p(Y|X)}{q(Y|X)} =
    \kldiv{p}{q} =
    \expect_{Y \sim p} \kldiv{p(X|Y)}{q(X|Y)} .
  \end{equation}
\end{lemma}
\begin{proof}
  The chain rule for KL divergence yields the equations
  \begin{align*}
    \kldiv{p}{q} & = \kldiv{p(X)}{q(X)} + \expect_{X \sim p}
      \kldiv{p(Y|X)}{q(Y|X)} \\
      & = \kldiv{p(Y)}{q(Y)} + \expect_{Y \sim p}
      \kldiv{p(X|Y)}{q(X|Y)} .
  \end{align*}
  The lemma now follows from the fact that the KL divergence
  of two identical distributions is zero.
\end{proof}
\begin{lemma}
  \label{lem:kl-dataproc}
  Suppose $\Omega, \Sigma$ are finite sets and $f : \Omega \to \Sigma$
  is a function. For two probability distributions $p,q$ on $\Omega$,
  let $p^f$ and $q^f$ denote the distributions of $f(\omega)$ when
  $\omega$ is sampled from $p$ or from $q$, respectively. Then
  \begin{equation} \label{eq:kl-dataproc}
    \kldiv{p}{q} \ge \kldiv{p^f}{q^f} .
  \end{equation}
\end{lemma}
\begin{proof}
  Let $\Gamma \subset \Omega \times \Sigma$ denote the graph of $f$,
  i.e. $\Gamma = \{ (\omega, f(\omega)) \mid \omega \in \Omega \} . $
  Let $\tilde{p}, \, \tilde{q}$ denote the distributions of
  $(\omega,f(\omega))$ when $\omega$ is sampled from $p$ or from
  $q$, respectively. The function $\omega \mapsto (\omega,f(\omega))$
  is a probability-preserving bijection between the probability
  spaces $(\Omega,p)$ and $(\Gamma,\tilde{p})$ and also between
  the probability spaces $(\Omega,q)$ and $(\Gamma,\tilde{q})$.
  Consequently, $\kldiv{p}{q} = \kldiv{\tilde{p}}{\tilde{q}} .$
  Now, using the chain rule and the non-negativity of KL divergence
  (\Cref{lem:kl-chainrule,lem:kl-nonneg}),
  \[
    \kldiv{\tilde{p}}{\tilde{q}} =
    \kldiv{p^f}{q^f} + \expect_{f(\omega) \sim p^f}
    \kldiv{p(\omega | f(\omega))}{q(\omega | f(\omega))}
    \ge
    \kldiv{p^f}{q^f} .
  \]
\end{proof}

Finally, we shall make use of two lemmas bounding the KL divergence
of Bernoulli distributions. The first is an upper bound on
$\kldiv{p}{q}$ when neither $q(0)$ nor $q(1)$ is close to zero.
The second is a lower bound on $\kldiv{p}{q}$ when
$q(0)$ is close to zero and $p(0)$ is far from zero.

\begin{lemma} \label{lem:kl-quadratic}
  If $p,q$ are two distributions on $\{0,1\}$ such
  that $q(0),q(1) > 0$, then
  \begin{equation} \label{eq:kl-quadratic}
    \kldiv{p}{q} \le
    (p(0) - q(0))^2 \left( \frac{1}{q(0)} + \frac{1}{q(1)} \right) .
  \end{equation}
\end{lemma}
\begin{proof}
  Using the inequality $\ln(1+x) \le x$ and the
  fact that $p(0)-q(0) = -(p(1) - q(1))$, we find that
  \begin{align*}
    \kldiv{p}{q} & =
    p(0) \ln \left( \frac{p(0)}{q(0)} \right) +
    p(1) \ln \left( \frac{p(1)}{q(1)} \right) \\
    & =
    p(0) \ln \left( 1 + \frac{p(0)-q(0)}{q(0)} \right) +
    p(1) \ln \left( 1 + \frac{p(1) - q(1)}{q(1)} \right) \\
    & \le
    p(0) \cdot \frac{p(0)-q(0)}{q(0)} +
    p(1) \cdot \frac{p(1)-q(1)}{q(1)} \\
    & =
    p(0) \cdot \frac{p(0)-q(0)}{q(0)} -
    p(1) \cdot \frac{p(0)-q(0)}{q(1)} \\
    & =
    (p(0)-q(0)) \cdot \left( \frac{p(0)}{q(0)} - \frac{p(1)}{q(1)} \right) \\
    & =
    (p(0)-q(0)) \cdot \left( 1 + \frac{p(0)-q(0)}{q(0)} -
    1 - \frac{p(1)-q(1)}{q(1)} \right) \\
    & =
    (p(0)-q(0)) \cdot \left( \frac{p(0)-q(0)}{q(0)} + \frac{p(0)-q(0)}{q(1)}
    \right) \\
    & =
    (p(0)-q(0))^2 \cdot \left( \frac{1}{q(0)} + \frac{1}{q(1)} \right) .
  \end{align*}
\end{proof}

\begin{lemma} \label{lem:kl-logarithmic}
  If $p,q$ are distributions on $\{0,1\}$ such that
  $q(1) \le \frac12 \le p(1)$ then
  \begin{equation} \label{eq:kl-logarithmic}
    \kldiv{p}{q} \ge \frac12 \ln \left( \frac{1}{4 q(1)} \right) .
  \end{equation}
\end{lemma}
\begin{proof}
  Let $x = p(0)$.
  We begin by calculating the partial derivative of
  $\kldiv{p}{q}$ with respect to $x$.
  \begin{align*}
    \frac{\partial}{\partial x} \kldiv{p}{q} & =
    \frac{\partial}{\partial x} \left[
      x \ln(x) - x \ln(q(0)) +
      (1-x) \ln(1-x) - (1-x) \ln(q(1))
      \right] \\
    & = ln(x) + 1 - \ln(q(0)) - \ln(1-x) - 1 + \ln(q(1)) \\
    & = \ln \left( \frac{x q(1)}{(1-x) q(0)} \right) \\
    & = \ln \left( \frac{x - x q(0)}{q(0) - x q(0)} \right) .
  \end{align*}
  The partial derivative is negative when $x < q(0)$
  and positive when $x > q(0)$. Since we are assuming
  $p(0) \le \frac12 \le q(0)$, as $p(0)$ varies over
  the range $[0,\frac12]$ the minimum of $\kldiv{p}{q}$
  occurs at $p(0)=\frac12$, when
  \[
  \kldiv{p}{q} = \frac12 \ln \left( \frac{1}{2 q(0)} \right)
  + \frac12 \ln \left(\frac{1}{2 q(1)} \right)
  = \frac12 \ln \left( \frac{1}{4 q(0) q(1)} \right)
  \ge \frac12 \ln \left( \frac{1}{4 q(1)} \right) .
  \]
\end{proof}

\begin{lemma} \label{lem:kl-product}
  Suppose $p$ and $q$ are product distributions on $k$-tuples
  $X=(X_1,\ldots,X_k)$ and that there exists a $k$-tuple
  $x^* = (x^*_1,x^*_2,\ldots,x^*_k)$ and a number $m \ge 2$
  such that
  $p(X=x^*) \ge \frac12$ while $q(X_i = x^*_i) \le \frac1m$
  for all $i$. Then
  \begin{equation} \label{eq:kl-product}
    \kldiv{p}{q} \ge \frac{k}{2} \ln \left( \frac{m}{4} \right)
    \quad \mbox{and} \quad
    \kldiv{q}{p} \ge \frac{k}{2} \ln \left( \frac{k}{3} \right) .
  \end{equation}
\end{lemma}
\begin{proof}
  From \Cref{lem:kl-additive} we know that
  $\kldiv{p}{q} = \sum_{i=1}^k \kldiv{p(X_i)}{q(X_i)}$
  and $\kldiv{q}{p} = \sum_{i=1}^k \kldiv{q(X_i)}{p(X_i)}$.
  For convenience let $p_i$ denote the distribution
  $p(X_i)$ and let $q_i$ denote the distribution
  $q(X_i)$. If we define $f(x_i)$ to be 1 if $x_i = x^*_i$
  and 0 if $x_i \neq x^*_i$, then $p_i^{f}(1) = p(X_i = x^*_i) \ge
  p(X = x^*) \ge \frac12$ and $q_i^f(1) = q(X_i = x^*_i) \le \frac1m \le \frac12.$
  By \Cref{lem:kl-dataproc,lem:kl-logarithmic},
  \[
    \kldiv{p_i}{q_i} \ge
    \kldiv{p_i^f}{q_i^f} \ge \frac12 \ln \left( \frac{1}{4 q_i^f(1)} \right)
    \ge \frac12 \ln \left( \frac{m}{4} \right) .
  \]
  Summing over $i$ we obtain the bound
  $ \kldiv{p}{q} \ge \frac{k}{2} \ln \left( \frac{m}{4} \right) .$

  To prove the lower bound on $\kldiv{q}{p},$
  first define $z_i = p(X_i \neq x_i^*).$ Since
  $p(X =  x^*) \ge \frac12$ we know that
  $\prod_{i=1}^k (1-z_i) \ge \frac12.$
  Using the AM-GM inequality,
  this implies
  \begin{align*}
    \frac1k \sum_{i=1}^k (1-z_i) & \ge (1/2)^{1/k} =
      e^{-\ln(2)/k} > 1 - \frac{\ln 2}{k} \\
    \frac1k \sum_{i=1}^k z_i & < \frac{\ln 2}{k} .
  \end{align*}
  Another application of the AM-GM inequality leads to
  \begin{align*}
    \prod_{i=1}^k z_i & < \left( \frac{\ln 2}{k} \right)^{k} \\
    \sum_{i=1}^k \ln(z_i) & < k \ln \left( \frac{\ln(2)}{k} \right)
      < k \ln \left( \frac{3}{4k} \right) \\
    \sum_{i=1}^k \ln(4 z_i) & < k \ln \left( \frac{3}{k} \right) .
  \end{align*}
  Now, we proceed similarly to the previous paragraph.
  Let $g(X_i) = 1$ if $X_i \neq x_i^*$ and $g(X_i)=0$
  if $X_i = x_i^*$. We have
  $p_i^g(1) = p_i(X_i \neq x_i^*) = z_i$,
  which is less than or equal to $\frac12$
  since the inequality $\prod_{i=1}^k (1-z_i) \ge \frac12$
  implies $1-z_i \ge \frac12$ for each $i$.
  Meanwhile, $q_i^g(1) = q_i(X_i \neq x_i^*) =
  1 - q_i(X_i = x_i^*) \ge 1 - \frac1m \ge \frac12.$
  By \Cref{lem:kl-dataproc,lem:kl-logarithmic},
  \[
    \kldiv{q_i}{p_i} \ge
    \kldiv{q_i^f}{p_i^f} \ge \frac12 \ln \left( \frac{1}{4 p_i^f(1)} \right)
    = - \frac12 \ln \left( 4 z_i \right) .
  \]
  Summing over $i$, we obtain
  \[
    \kldiv{q}{p} \ge
    - \frac12 \sum_{i=1}^k \ln(4 z_i) >
    \frac{k}{2} \ln \left( \frac{k}{3} \right) .
  \]
\end{proof}

\subsection{Completing the proof} \label{sec:lb-finish}
\newcommand{\pind}{{\mathring{p}}}
\newcommand{\femp}{{F_{\mathrm{emp}}}}
\newcommand{\thmle}{\theta_{\mathrm{MLE}}}
\newcommand{\thalg}{\hat{\theta}}

In this section we complete the proof of
\Cref{thm:low-bd}. To this end, suppose
that $n > 1/\eps$ and that we are given an
algorithm that uses $T$ threshold queries
to produce a CDF estimate that is $\eps$-accurate
with probability at least $\frac34$. We aim to
prove that
$T \ge \Omega(k \log(n) / \eps^2) = \Omega(\log(n)/\eps^3).$

Let $H = ((q_1,y_1),\ldots,(q_T,y_T))$ denote the
random history of queries and responses obtained
when running the algorithm on a (potentially random)
sequence $x_1,\ldots,x_T$. We will consider two
joint distributions $p,\pind$ over pairs $(\theta,H)$
consisting of a parameter vector $\theta$ and
history $H$. Distribution $p$ is the distribution
obtained by sampling parameter vector $\theta \in [m]^k$
uniformly at random, then drawing $T$ independent samples
$x_1,\ldots,x_T$ from $D_{\theta}$, and finally running
the algorithm on input sequence $x_1,\ldots,x_T$ to generate
a history, $H$. Distribution $\pind$ is the product distribution
$p(\theta) \times p(H)$. In other words, a random sample
from $\pind$ is defined by sampling a random history $H$
from the marginal distribution $p(H)$, and independently
drawing a uniformly random parameter vector $\theta \in [m]^k$
that bears no relation to $H$.

Since $p$ and $\pind$ have identical marginals,
\Cref{lem:kl-bayesian} tells us that
\begin{align} \label{eq:klbayes.pq}
  \expect_{\theta} \kldiv{p(H|\theta)}{\pind(H|\theta)}
  & =
  \kldiv{p}{\pind} =
  \expect_H \kldiv{p(\theta|H)}{\pind(\theta|H)} \\
  \label{eq:klbayes.qp}
  \expect_{\theta} \kldiv{\pind(H|\theta)}{p(H|\theta)}
  & =
  \kldiv{p}{\pind} =
  \expect_H \kldiv{\pind(\theta|H)}{p(\theta|H)}
\end{align}

The next lemma furnishes an upper bound on the quantities
appearing on the left sides of~\eqref{eq:klbayes.pq}
and~\eqref{eq:klbayes.qp}.

\begin{lemma} \label{lem:klbayes-ub}
  For any $\theta \in [m]^k$,
  \begin{equation} \label{eq:klbayes-ub}
     \kldiv{p(H|\theta)}{\pind(H|\theta)} \le 48 \eps^2 T
     \quad \mbox{and} \quad
     \kldiv{\pind(H|\theta)}{p(H|\theta)} \le 48 \eps^2 T
  \end{equation}
\end{lemma}
\begin{proof}
We use the chain rule for KL divergence, \Cref{lem:kl-chainrule}.
For $s=1,2,\ldots,T$ let $H_s$ denote the random variable
$((q_1,y_1),\ldots,(q_s,y_s))$ consisting of the first $s$
queries and responses.
\begin{equation} \label{eq:klbub.1}
  \kldiv{p(H|\theta)}{\pind(H|\theta)} =
  \sum_{s=1}^T \kldiv{p((q_s,y_s) | H_{s-1},\theta)}{\pind((q_s,y_s) | H_{s-1},\theta)}
\end{equation}
A second application of the chain rule for KL divergence
allows us to break down each term of the sum even further.
\begin{align} \label{eq:klbub.2}
  \kldiv{p((q_s,y_s) | H_{s-1},\theta)}{\pind((q_s,y_s) | H_{s-1},\theta)} & =
  \kldiv{p(q_s | H_{s-1},\theta)}{\pind(q_s | H_{s-1},\theta)} \\
  \nonumber & \; \; +
  \kldiv{p(y_s | q_s,H_{s-1},\theta)}{\pind(y_s | q_s,H_{s-1},\theta)}
\end{align}
Since the algorithm selects $q_s$ based on $H_{s-1}$ only,
the conditional distributions $p(q_s | H_{s-1},\theta)$ and
$\pind(q_s | H_{s-1},\theta)$ are identical, hence the first
term on the right side of~\eqref{eq:klbub.2} is zero:
\begin{equation} \label{eq:klbub.2.5}
  \kldiv{p(q_s | H_{s-1},\theta)}{\pind(q_s | H_{s-1},\theta)} = 0
\end{equation}
As for the second term, to simplify notation we will
denote $p(y_s | q_s,H_{s-1},\theta)$ and $\pind(y_s | q_s, H_{s-1}, \theta)$
by $p_s$ and $\pind_s$, respectively. Then
$p_s(0) = 1-F_{\theta}(q_s)$,
whereas
$\pind_s(0) = \expect[1 - F_{\theta'}(q_s)]$
with the expectation on the right side being computed by
sampling $\theta'$ from the conditional distribution $p(\theta | H_{s-1})$.
\Cref{lem:ftheta-close} tells us that $|F_{\theta}(q_s) - F_{\theta'}(q_s)|
\le 3 \eps$ for every $\theta'$, so
\begin{equation} \label{eq:klbub.3}
  | p_s(0) - \pind_s(0) | \le 3 \eps.
\end{equation}
If $q_s=n$ then $y_s$ is deterministically equal to
$1$ under both distributions, $p$ and $\pind$, so
$\kldiv{p_s}{\pind_s} = 0$
when $q_s=n$. Otherwise,
$ \pind_s(0) $
belongs to the interval $\left[ \frac14, \frac34 \right]$
and $\pind_s(1) = 1 - \pind_s(0)$,
so
\begin{equation} \label{eq:klbub.3.5}
  \frac{1}{\pind_s(0)} +
  \frac{1}{\pind_s(1)}  \le
   \frac{1}{1/4} + \frac{1}{3/4} = \frac{16}{3} .
\end{equation}
Now, applying \Cref{lem:kl-quadratic},
\begin{equation} \label{eq:klbub.4}
  \kldiv{p_s}{\pind_s}
  \le
  (p_s(0) - \pind_s(0))^2
  \cdot \frac{16}{3}
  \le
  \frac{16}{3} (3 \eps)^2 = 48 \eps^2 .
\end{equation}
The inequality $\kldiv{p(H|\theta)}{\pind(H|\theta)} \le 48 \eps^2 T$
follows by combining
lines~\eqref{eq:klbub.1},\eqref{eq:klbub.2},\eqref{eq:klbub.2.5},\eqref{eq:klbub.4}.
The second inequality in the statement of the lemma
follows by an identical argument with the roles of $p$ and $\pind$
reversed.
\end{proof}

The proof of \Cref{thm:low-bd} concludes as follows.
\begin{proof}[Proof of \Cref{thm:low-bd}]
  Theorem 6 of \citet{meister} already proves
  there is a universal constant $c > 0$ such that
  $T \ge c n \log(n) / \eps^2$ when $\eps \le 1/(n+1).$
  This has two consequences for our proof. First, to
  complete the proof we only need to consider the case
  that $\eps > 1/(n+1)$, in which case the theorem
  asserts a lower bound of the form $T = \Omega(\log(n) / \eps^3).$
  Second, if we define
  $n' = \left\lfloor \frac{1}{\eps} \right\rfloor - 1$
  then $\eps \le 1/(n'+1)$, so Theorem 6 of \citet{meister}
  proves that the CDF estimation problem for distributions on
  $[n']$ requires at least
  \begin{equation} \label{eq:tlb.1}
    c n' \log(n') / \eps^2 \ge c' \ln(1/\eps) / \eps^3
  \end{equation}
  samples. (Here, $c' > 0$ is another univeral constant.)
  Since CDF estimation on $[n]$ generalizes
  CDF estimation on $[n']$, the sample complexity
  of CDF estimation on $[n]$ is also bounded below
  by the right side of~\eqref{eq:tlb.1}:
  \begin{equation} \label{eq:tlb.2}
    T \ge c' \ln(1/\eps) / \eps^3 .
  \end{equation}
  Recall from \Cref{sec:intro} that CDF estimation
  generalizes binary search, so $T \ge \lfloor \log(n) \rfloor
  \ge \frac12 \log(n)$.
  If $\eps \ge \min \{c', 1/256 \}$ then
  $\frac12 \log(n) \ge \frac12 \cdot (\min \{c',1/256\})^{-3} \cdot
  \log(n)/\eps^3$, so $T = \Omega(\log(n)/\eps^3)$ as claimed.
  Thus, for the remainder of the proof we may assume
  \begin{equation} \label{eq:tlb.3}
    \eps < \min \left\{ c', \frac{1}{256} \right\}.
  \end{equation}

  Let $\femp$ denote the empirical distribution of
  the $T$ samples $x_1,\ldots,x_T$.
  Under distribution $p$, these samples are i.i.d.~draws
  from $D_{\theta},$ so by
  the Dvoketzky-Kiefer-Wolfowitz Inequality \citep{dkw}
  \begin{equation} \label{eq:tlb.4}
    p \left( \| \femp - F_{\theta} \|_{\infty} \ge \frac{\eps}{2} \right)
    \le
    2 e^{- T \eps^2 / 2} .
  \end{equation}
  Substituting the lower bound for $T$ in \eqref{eq:tlb.2}
  and the upper bound for $\eps$ in \eqref{eq:tlb.3} we find
  that
  \begin{equation} \label{eq:tlb.5}
    2 e^{-T \eps^2 / 2} \le
    2 e^{- c' \ln(1/\eps) / (2 \eps)} \le
    2 e^{-\ln(1/\eps) / 2} \le
    2 e^{-\ln(256) / 2} = \frac18 .
  \end{equation}
  If $\hat{F}$ denotes the CDF estimate produced
  by our algorithm, then the algorithm's probabilistic
  approximate correctness guarantee asserts that
  \begin{equation} \label{eq:tlb.6}
    p \left( \| \hat{F} - \femp \|_{\infty} > \eps \right)
    \le \frac14 .
  \end{equation}
  Combining inequalities~\eqref{eq:tlb.3},\eqref{eq:tlb.4},\eqref{eq:tlb.5}
  and using the union bound and triangle inequality, we have
  \begin{align*}
    p \left( \| \femp - F_{\theta} \|_{\infty} \ge \frac{\eps}{2}
    \mbox{ or } \| \hat{F} - \femp \|_{\infty} > \eps \right)
    & \le \frac38 \\
    p \left( \| \femp - F_{\theta} \|_{\infty} < \frac{\eps}{2}
    \mbox{ and } \| \hat{F} - \femp \|_{\infty} \le \eps \right)
    & \ge \frac58 \\
    p \left( \| \hat{F} - F_{\theta} \|_{\infty} < \frac{3 \eps}{2}
    \right) & \ge \frac58 .
  \end{align*}
  Now let us define two random variables
  $\thalg, \, \thmle$ as follows:
  $\thalg$ is the value of $\theta$ whose
  associated CDF $F_{\thalg}$ is nearest
  to $\femp$ in $L_{\infty},$ while
  $\thmle$ is the value of $\theta$ whose
  conditional probability
  $p(\theta = \thmle \mid H)$ is greatest.
  From \Cref{lem:ftheta-close} we know that
  $\theta = \thalg$ whenever
  $\| \hat{F} - F_{\theta} \|_{\infty} < \frac{3 \eps}{2}$,
  so $p(\theta = \thalg) \ge \frac58.$
  By the definition of $\thmle$, we have
  \begin{equation} \label{eq:tlb.7}
    p(\theta = \thmle | H) \ge p(\theta = \thalg | H)
  \end{equation}
  for all $H$. Take the expectation of both sides
  of~\eqref{eq:tlb.7} with respect to $H$ and use
  the law of iterated expectation to deduce
  \begin{equation} \label{eq:tlb.8}
    p(\theta = \thmle) \ge p(\theta = \thalg) = \frac58 .
  \end{equation}
  Let us define a {\em good history} to be a history
  $H$ such that $p(\theta = \thmle | H) \ge \frac12.$
  We have
  \begin{equation} \label{eq:tlb.9}
    \frac58 \le p(\theta = \thmle) \le
    p(H \mbox{ is good}) \cdot 1 +
    p(H \mbox{ is not good}) \cdot \frac12 =
    \frac12 + \frac12 p(H \mbox{ is good}),
  \end{equation}
  so $p(H \mbox{ is good}) \ge \frac14.$

  When the history $H$ is good, it means that
  $p(\theta = \thmle|H) \ge \frac12$  whereas
  $q(\theta|H)$ is the uniform distribution over
  $[m]^k$ so $q(\theta_i = (\thmle)_i | H) = \frac1m$
  for all $i \in [k]$. The conditions for
  \Cref{lem:kl-product} are satisfied, so
  we may conclude that the inequalities
  $ \kldiv{p(\theta|H)}{q(\theta|H)} \ge \frac{k}{2} \ln
  \left( \frac{m}{4} \right)$
  and
  $ \kldiv{q(\theta|H)}{p(\theta|H)} \ge \frac{k}{2} \ln
  \left( \frac{k}{3} \right) $
  hold when $H$ is good. Since the probability
  that $H$ is good is at least $\frac14$, we have
  \begin{align} \label{eq:tlb.10}
    \expect_H \kldiv{p(\theta|H)}{q(\theta|H)} & \ge \frac{k}{8} \ln
    \left( \frac{m}{4} \right) \\
    \label{eq:tlb.11}
    \expect_H \kldiv{q(\theta|H)}{p(\theta|H)} & \ge \frac{k}{8} \ln
    \left( \frac{k}{3} \right) .
  \end{align}
  Using \Cref{lem:klbayes-ub} together with Equations~\eqref{eq:klbayes.pq} and~\eqref{eq:klbayes.qp}, we find that
  \begin{align} \label{eq:tlb.11}
    48 T \eps^2 & \ge \expect_{\theta} \kldiv{p(H|\theta)}{q(H|\theta)}
    = \expect_H \kldiv{p(\theta|H)}{q(\theta|H)} \ge
    \frac{k}{8} \ln
    \left( \frac{m}{4} \right) \\
    \label{eq:tlb.12}
    48 T \eps^2 & \ge \expect_{\theta} \kldiv{q(H|\theta)}{p(H|\theta)}
    = \expect_H \kldiv{q(\theta|H)}{p(\theta|H)} \ge
    \frac{k}{8} \ln
    \left( \frac{k}{3} \right) .
  \end{align}
  Summing inequalities~\eqref{eq:tlb.11} and~\eqref{eq:tlb.12},
  we obtain
  \begin{equation} \label{eq:tlb.13}
    96 T \eps^2 \ge \frac{k}{8} \ln \left( \frac{km}{12} \right) .
  \end{equation}
  Recalling that $k = \frac{1}{6 \eps}$ and that $km = n$, we find
  that $T > \frac{1}{4800 \eps^3} \ln \left( \frac{n}{12} \right)$,
  which concludes the proof that $T = \Omega(\log(n) / \eps^3).$
\end{proof}

\section{Proof of Theorem~\ref{thm:import-wts}}
\label{sec:import-wts-proof}

\begin{proof}
  In the proof, we will use $\E_t[\cdots]$ as a notation for
  the conditional expectation of random variables, conditioning on the
  history of the first $t-1$ rounds and on the adversary's choice of $\vgedit{v}_t$. We make a similar notation change for $\var [\cdots]$.
% \rdkdelete{We assume that $y_{1,t} \leq \ldots \leq y_{k,t}$ for all $t$. \footnote{This is a reasonable assumption to make because if there was some $i$ such that $y_{i,t} \ge y_{i+1, t}$, we can set $y_{i,t}$ to $y_{i+1, t}$ and still satisfy the guarantee that $\forall i \  |y_{i,t} - F_t(q_{i,t})| \leq \delta$. This is because $F_t$ is a monotone function.}}
We note that $\E_t [\hat{u}_t] = u_t$ and $\E_t [\hat{\ell}_t] = \ell_t$. This follows from the fact that $u_t(i) = k \cdot u_t(i)$ with probability $\frac{1}{k}$ and 0 otherwise. The argument follows similarly for $\ell_t$. Thus, $\E_t [\hat{d}_t] = d_t$.
Intuitively, Algorithm~\ref{algo:import-wt} works because even though we don't observe $d_t$, we can still obtain an unbiased estimate of $d_t$.

The weights $w_{i,t}$ and coefficients $a_{i,t}$ in Algorithm~\ref{algo:mult-wt} evolve according to the update equations of the standard EXP3 algorithm \rdkedit{of \citet{auer}. By the analysis of that algorithm,}
\rdkcomment{Would be nice to reference an actual theorem
from that paper here, so that readers could easily find the bound
we're using from the Exp3 paper.}
%\pcocomment{add citation here}, and according to the analysis of that algorithm, we have the inequality
    \begin{equation} \label{eq:exp3}
        \sum_{t=1}^T \sum_{i=1}^n a_i \hat{d}_t(i) \ge
        \max_{i \in [n]} \left\{ \sum_{t=1}^T \hat{d}_t(i) \right\}
        - \frac{\ln n}{\eta} - \eta \sum_{t=1}^T \sum_{j=1}^n a_i \hat{d}^2_t(i).
    \end{equation}
    \rdkcomment{In the foregoing displayed equation, and almost every other one on this page, I revised the size of curly braces from $\{ \sum_{t=1}^T \cdots \}$
    to $\left\{ \sum_{t=1}^t \cdots \right\}$.}
Taking expectations, we get that

\begin{equation}
        \sum_{t=1}^T \sum_{i=1}^n a_i d_t(i) \ge
         \max_{i \in [n]} \left\{ \sum_{t=1}^T d_t(i) \right\}
        - \frac{\ln n}{\eta} - \eta \sum_{t=1}^T \sum_{j=1}^n a_i \E [\hat{d}^2_t(i)].
\end{equation}
\rdkedit{%
To bound the last term on the right side, we make use of
the law of iterated expectation: $\E[\hat{d}^2_t(i)] = \E[\E_t[\hat{d}^2_t(i)]].$
Now, if $i = q_{j,t}$ for some $j$ then $\hat{u}_t(i) = \hat{\ell}_t(i)$ so
$\hat{d}_t(i) = 0$. On the other hand, if $q_{j,t} < i < q_{j+1,t}$ then
$\hat{d}_t(i) = k \cdot u_t(i)$ with probability $1/k$,
$\hat{d}_t(i) = - k \cdot \ell_t(i)$ with probability $1/k$,
and otherwise $\hat{d}_t(i) = 0$. Thus,
\begin{equation}
  \E_t[\hat{d}^2_t(i)] = \frac1k \left( k^2 u^2_t(i) + k^2 \ell^2_t(i) \right)
  \le \frac1k \left( k^2 + k^2 \right) = 2k .
\end{equation}
Applying the law of iterated expectation, we have
$\E[\hat{d}^2_t(i)] \le 2k.$
}
% $\E_t [\hat{d}^2_t(i)] = \E_t [\hat{u}^2_t(i)] + \E_t [\hat{\ell}^2_t(i)] - 2\E [u_t(i) \ell_t (i)]$. Since $\hat{u}^2_t(i) = k^2 \cdot u^2_t(i)$ with probability $\frac{1}{k}$ and 0 otherwise, $\E [\hat{u}^2_t(i)] = k u^2_t(i) \leq k$. By a similar argument, $\E [\hat{\ell}^2_t(i)] = k \ell^2_t(i) \leq k$ and thus, $\E [\hat{d}^2_t(i)] \leq 2k$.
This, together with the fact that $\sum_{j=1}^n a_i \leq 1$, allows us to simplify the equation to
\begin{equation}
        \sum_{t=1}^T \sum_{i=1}^n a_i d_t(i) \ge
         \max_{i \in [n]} \left\{ \sum_{t=1}^T d_t(i) \right\}
        - \frac{\ln n}{\eta} - 2 \eta T k
\end{equation}
From the proof of \Cref{prop:approachable} we know that for all $t$, \rdkedit{$\sum_{i=1}^n a_i d_t(i) \le  \frac{1}{k+1}$}. Substituting this bound
\rdkedit{%
\begin{equation}
       \frac{T}{k+1} \ge
         \max_{i \in [n]} \left\{ \sum_{t=1}^T d_t(i) \right\}
        - \frac{\ln n}{\eta} - 2 \eta k T
\end{equation}
\begin{equation}
       \frac{T}{k+1} + \frac{\ln n}{\eta} + 2\eta k T \ge
         \max_{i \in [n]} \left\{ \sum_{t=1}^T d_t(i) \right\}
\end{equation}
}
Recalling that \rdkedit{$k = \frac{2}{\eps}, \, \eta = \frac{\eps^2}{16} = \frac{\eps}{8k}, T \ge \frac{64 \ln(n)}{\eps^3}$, we find that
    $\frac{T}{k+1} \le \frac{\eps T}{2}, \,
    \frac{\ln n}{\eta} \le \frac{\eps T}{4}, \,
    2 \eta k T \le \frac{\eps T}{4}.$ Hence
    \begin{align*}
        \frac{\varepsilon}{2} T +
        \frac{\varepsilon}{4} T  +
        \frac{\varepsilon}{4} T  & \ge
        \max_{i \in [n]} \left\{ \sum_{t=1}^T d_t(i) \right\} \\
        \eps & \ge
        \max_{i \in [n]} \left\{ \tfrac 1T \sum_{t=1}^T d_t(i) \right\} .
    \end{align*}
    }%
The right side of the last inequality is equal to the width of the interval $\left[  \frac1T \sum_{t=1}^T \ell_t(i), \; \frac1T \sum_{t=1}^T u_t(i) \right]$. That interval is guaranteed to contain $\frac1T \sum_{t=1}^T \vgedit{v}_t(i)$, and its midpoint is $G_T(i)$, so we are assured that \rdkedit{$|G_T(i) - \frac1T \sum_{t=1}^T \vgedit{v}_t(i)| \le \frac{\eps}{2}$.} However, because the algorithm only observes estimates $\hat{u}_t$ and $\hat{\ell}_t$, it doesn't know the actual values of $\frac1T \sum_{t=1}^T \ell_t(i)$ and $\frac1T \sum_{t=1}^T u_t(i)$. We will now rely on concentration results to show that after $T \ge \frac{64 \ln n}{\varepsilon^3}$, with high probability, the estimates $\frac1T \sum_{t=1}^T \hat{\ell}_t(i)$ and $\frac1T \sum_{t=1}^T \hat{u}_t(i)$ will be close to the true values $\frac1T \sum_{t=1}^T \ell_t(i)$ and $\frac1T \sum_{t=1}^T u_t(i)$. Since the estimates, $\hat{\ell}_t$ and $\hat{u}_t$, are correct in expectation and $\E [\hat{\ell}^2_t(i)] = k \ell^2_t(i)$ (also $\E [\hat{u}^2_t(i)] = k u^2_t(i)$), we have
% \begin{equation}
%     \E \left[ \frac1T \sum_{t=1}^T \hat{\ell}_t(i) - \frac1T \sum_{t=1}^T \ell_t(i) \right]^2 = \frac{k-1}{t} \left( \frac1T \sum_{t=1}^T \ell_t(i) \right) \leq \frac{k}{t}
% \end{equation}
\begin{equation}
    \var \left( \frac1T \sum_{t=1}^T \hat{\ell}_t(i) \right) = \frac{1}{T^2} \sum_{t=1}^T \var  \left( \hat{\ell}_t(i) \right) =  \frac{1}{T^2} \sum_{t=1}^T (k-1) \ell_t^2 (i) \leq \frac{k}{T}
\end{equation}
for each $i \in [n]$. Using the martingale variant of Bernstein's inequality proved in \cite{freedman}, we have
\begin{equation}
    \prob \left[ \left| \frac1T \sum_{t=1}^T \hat{\ell}_t(i) - \frac1T \sum_{t=1}^T \ell_t(i) \right| \geq \frac{\varepsilon}{2} \right] \leq 2\exp \left(-\frac{T^2 (\varepsilon/2)^2}{2(k-1)(\frac1T \sum_{t=1}^T \ell_t(i)) + \frac{2}{3} k T (\varepsilon/2)} \right) \leq 2\exp \left(- \frac{T (\varepsilon/2)^2}{3k} \right)
\end{equation}
which is less than $\frac{1}{4n}$ since $T \ge \frac{64 \ln n}{\varepsilon^3}$.

\noindent Following similar analysis for $u_t(i)$, we get that $\prob \left[ \left| \frac1T \sum_{t=1}^T \hat{u}_t(i) - \frac1T \sum_{t=1}^T u_t(i) \right| \geq \frac{\varepsilon}{2} \right]$ is less than $\frac{1}{4n}$. As a result, with high probability, the interval $\left[  \frac1T \sum_{t=1}^T \hat{\ell}_t(i), \frac1T \sum_{t=1}^T \hat{u}_t(i) \right]$ contains $\frac1T \sum_{t=1}^T \vgedit{v}_t(i)$ and is $\frac{\varepsilon}{2}$ close to the midpoint of the true interval $\left[  \frac1T \sum_{t=1}^T \ell_t(i), \frac1T \sum_{t=1}^T u_t(i) \right]$. Thus, with high probability,
\begin{equation}
    \left| \hat{G}_T(i) - \frac1T \sum_{t=1}^T \vgedit{v}_t(i) \right| \leq \left| \hat{G}_T(i) - G_T(i) \right| + \left| G_T(i) - \frac1T \sum_{t=1}^T \vgedit{v}_t(i) \right| \leq \frac{\varepsilon}{2} + \frac{\varepsilon}{2} = \varepsilon \ \text{as desired.}
\end{equation}
\end{proof}

\section{Elementary Algorithms for Threshold Query Model}\label{sec:elementary}

\begin{lemma}\label{pf:deterministic}
     No deterministic algorithm can obtain accuracy $\eps$ with a query budget less than $\frac{1}{2 \varepsilon} - 1$
\end{lemma}
\begin{proof}
    Suppose for sake of contradiction that an algorithm $\mathcal{A}$ obtains accuracy $\eps$ with a query budget of $k = \frac{1}{2\eps} - 2$. Suppose $n \geq k+1$. After $T$ timesteps, there exists a point, $p$, that has not been queried for at least $\frac{T}{k+1}$. If this was not the case, then all the points would have been queried for $n \cdot \frac{kT}{k+1} \geq (k+1) \cdot \frac{kT}{k+1} \geq kT$ (a contradiction). Let $G_T(p)$ be the algorithm's estimate at that point at time $T$ and let $F_T (p)$ be the average during timesteps when $p$ was queried. Since the algorithm doesn't know whether the function's value at $p$ was $0$ or $1$ during the $\frac{T}{k+1}$ timesteps, it follows that
    \begin{equation}
        |G(p) - F_T (p)| > \frac{1}{2} \left| \left(F_T (p) + \frac{1}{k+1} \right) - \left(F_T (p) + \frac{0}{k+1}\right) \right| > \frac{1}{2} \left(\frac{1}{k+1} \right) > \eps
    \end{equation}
\end{proof}

\begin{lemma}
    The pair $\left( k = (\frac{1}{\epsilon}+1) \sqrt{n}, T_0 = \frac{1}{\epsilon}\right)$ is achievable for monotone step functions. The algorithm behaves as follows: the algorithm always queries points $\{0, \lfloor \sqrt{n} \rfloor, 2\lfloor \sqrt{n} \rfloor, \ldots, n \}$ every timestep. The algorithm maintains the average $F$ of the step functions at these points. For any interval, $[m \sqrt{n}, (m+1) \sqrt{n}]$ such that $F[(m+1)\sqrt{n}] - F[m\sqrt{n}] \geq \epsilon$, the algorithm queries every point in the interval.
\end{lemma}
\begin{proof}
%We prove this by induction on $T$. Define the uncertainty gap of an interval to be We will show that after each timestep $t$, the uncertainty gap of every interval is either
%$|G - \frac{1}{T}\sum F_t|_\infty \leq \max \{\frac{1}{t}, \epsilon \}$. The base case $t=1$ follows from the fact that $F_1$ can be at most 1. Assume the statement is true after the $t$-th timestep. At timestep $t+1$, every interval $(m\sqrt{n}, (m+1)\sqrt{n})$ such that $F[(m+1)\sqrt{n}] - F[m\sqrt{n}] = \max \{\frac{1}{t}, \epsilon \}$ gets queried by the algorithm. Note that by IH, no interval has a gap of more than $\max \{\frac{1}{t}, \epsilon \}$. Thus,

%A stronger inductive hypothesis is to show that for every interval, the uncertainty gap is a multiple of $\frac{1}{t}$

This claim follows by observing that from timesteps 1 till $\frac{1}{\epsilon}$, for every interval $(m\sqrt{n}, (m+1)\sqrt{n})$, there is at most 1 timestep where we didn't know the value of every point in the interval.
\end{proof}

\begin{lemma}
    The pair $\left( k = (\frac{4}{\epsilon}+1) \sqrt{n}, T_0 = \frac{4}{\epsilon} \right)$ is robustly-achievable \footnote{In our original model, we assumed the algorithm observed $y_p$ such that $|y_p - F_{t+1}(p)| \leq \epsilon/4$ instead of directly receiving $F_{t+1}(p)$. This was done to make simulation easier but turns out to be unnecessary for the final algorithm.} for any monotone function. The algorithm behaves as follows: the algorithm always queries points $\{0, \lfloor \sqrt{n} \rfloor, 2\lfloor \sqrt{n} \rfloor, \ldots, n \}$ on every timestep. The algorithm maintains the average $\hat{F}$ of the step functions at these points. For any interval, $[m \sqrt{n}, (m+1) \sqrt{n}]$ such that $\hat{F}[(m+1)\sqrt{n}] - \hat{F}[m\sqrt{n}] \geq \epsilon/4$, the algorithm queries every point in the interval.
\end{lemma}
\begin{proof}
We prove this by induction on $T$. We will show that after each timestep $t$, $|\hat{F} - \frac{1}{t}\sum \vgedit{v}_t|_\infty \leq \frac{1}{t} + \frac{3\epsilon}{4} $. Assume the statement is true after the $t$-th timestep. At timestep $t+1$, every interval $(m\sqrt{n}, (m+1)\sqrt{n})$ such that $\hat{F}[(m+1)\sqrt{n}] - \hat{F}[m\sqrt{n}] \geq \epsilon/4$ gets queried by the algorithm. Note that by IH, no interval has a gap of more than $\frac{1}{t+1} + \frac{3\epsilon}{4}$ . Thus, after the $t+1$ timestep, for any point $p$ in a tracked interval (i.e one such that $\hat{F}[(m+1)\sqrt{n}] - \hat{F}[m\sqrt{n}] \geq \epsilon/4$),
\begin{align*}
    \left|\hat{F}^{t+1} (p) - \frac{1}{t+1}\sum_{s=0}^{t+1} \vgedit{v}_s (p) \right|\
    &\leq \left|\frac{1}{t+1} \left(t \cdot \hat{F}^{t} (p) + y_p  \right) - \frac{1}{t+1} \left( \sum_{s=0}^{t} \vgedit{v}_s (p) + \vgedit{v}_{t+1}(p) \right) \right|\ \\
    &\leq \left|\frac{1}{t+1} \left(t \cdot \hat{F}^{t} (p) - \sum_{s=0}^{t} \vgedit{v}_s (p) \right) + \frac{1}{t+1} \left(y_p   - \vgedit{v}_{t+1}(p) \right) \right|\ \\
\intertext{by inductive hypothesis}
    &< \left|\frac{t}{t+1} \left(\frac{1}{t} + \frac{3\epsilon}{4} \right) \right|
    + \left| \frac{1}{t+1} \left(y_p   - \vgedit{v}_{t+1}(p) \right) \right|\ \\
\intertext{by guarantee $|y_p - \vgedit{v}_{t+1}(p)| \leq \epsilon/4$ \vgcomment{just wanted to double check that it is correct to change this to lowercase f}} 
    &< \frac{1}{t+1} + \frac{t(3\epsilon/4)}{t+1}  +  \frac{\epsilon/4}{t+1} \\
    &<  \frac{1}{t+1} + \frac{3\epsilon}{4}
\end{align*}
For an untracked interval $[p_1, p_2]$, that is, one such that $\hat{F}[p_2] - \hat{F}[p_1] < \epsilon/4$, we have that after the $t+1$ timestep,
\begin{align*}
    \left|\hat{F}^{t+1} (p_2) - \hat{F}^{t+1} (p_1) \right|\
    &\leq \left|\frac{1}{t+1} \left(t \cdot \hat{F}^{t} (p_2) + y_{p_2}  \right) - \frac{1}{t+1} \left(t \cdot \hat{F}^{t} (p_1) + y_{p_1}  \right) \right|\ \\
    &\leq \left|\frac{t}{t+1} \left(\hat{F}^{t} (p_2) - \hat{F}^{t} (p_1) \right) + \frac{1}{t+1} \left(y_{p_2}  - y_{p_1} \right) \right|\ \\
\intertext{since the interval is untracked}
    &< \left|\frac{t}{t+1} \left(\epsilon/4 \right) + \frac{1}{t+1} \right|\ \\
    &< \frac{1}{t+1} + \frac{\epsilon}{4}
\end{align*}
% For a point $p$ in an untracked interval $[p_1,p_2]$ (i.e one such that $\hat{F}[p_2] - \hat{F}[p_1] < \epsilon/4$).
% \vgcomment{i think this sentence needs to be removed - was probably forgotten in editing} 

Since points $p_1$ and $p_2$ have been tracked from the start, it follows that $\left|\hat{F}^{t+1} (p_1) - \frac{1}{t+1}\sum_{s=0}^{t+1} \vgedit{v}_s (p_1) \right| \leq \epsilon/4$ and $\left|\hat{F}^{t+1} (p_2) - \frac{1}{t+1}\sum_{s=0}^{t+1} \vgedit{v}_s (p_2) \right| \leq \epsilon/4$. The $\epsilon/4$ comes from the noise assumption i.e algorithm observes $y$ such that $|y - \vgedit{v}_t| \leq \epsilon/2$ but does not observe $\vgedit{v}_t$ exactly. By monotonicity, we know that $\frac{1}{t+1}\sum_{s=0}^{t+1} \vgedit{v}_s (p_1) \leq \frac{1}{t+1}\sum_{s=0}^{t+1} \vgedit{v}_s (p)  \leq \frac{1}{t+1}\sum_{s=0}^{t+1} \vgedit{v}_s (p_2) $.
Thus, we have that for a point $p$ in an untracked interval $[p_1,p_2]$
\begin{align*}
    \left|\hat{F}^{t+1} (p) - \frac{1}{t+1}\sum_{s=0}^{t+1} \vgedit{v}_s (p) \right|\
    &\leq \left|\hat{F}^{t+1} (p_1) - \frac{1}{t+1}\sum_{s=0}^{t+1} \vgedit{v}_s (p_1) \right| + \left|\hat{F}^{t+1} (p_2) - \hat{F}^{t+1} (p_1) \right|\ \\
    &\quad + \left|\hat{F}^{t+1} (p_2) - \frac{1}{t+1}\sum_{s=0}^{t+1} \vgedit{v}_s (p_2) \right|\\
    &< \epsilon/4 + \epsilon/4 + \frac{1}{t+1} + \epsilon/4 \\
    &<  \frac{1}{t+1} + \frac{3\epsilon}{4}
\end{align*}
Thus, after $T \geq \frac{4}{\epsilon}$, it follows that $|\hat{F} - \frac{1}{T}\sum \vgedit{v}_T|_\infty \leq \frac{1}{T} + \frac{3\epsilon}{4} \leq \epsilon$.
\end{proof}

\begin{lemma}
    The pair $\left( k = \frac{\log n}{\epsilon^2}, T_0 = \frac{\log n}{\epsilon}\right)$ is achievable for monotone step functions. The algorithm behaves as follows: insert a new query point at the midpoint of the interval with the uncertainty of the current timestep. That is, suppose your query points were $q_1, q_2, \ldots, q_k$. There's an interval $(q_i, q_{i+1})$ where $\vgedit{v}_t[q_i] = 0$ and $\vgedit{v}_t[q_{i+1}] = 1$. Insert the new query point at $q = \lfloor (q_i + q_{i+1})/2 \rfloor$
\end{lemma}
\begin{proof}
    We prove this by induction on $T$. We will show that after every timestep $t$, for any point $p \in [p_1, p_2]$ where $p_1, p_2$ are adjacent active points, then $|\hat{F}[p] - \frac{1}{t}\sum \vgedit{v}_t[p]| \leq \frac{\log n - \log (p_2 - p_1)}{2t}$. Assume the statement is true after the $t$-th timestep. At timestep $t+1$, the adversary chooses a point $r$ to make the step for $\vgedit{v}_{t+1}$. Let $[r_1, r_2]$ be the adjacent active points that $r$ falls in. Since the algorithm will observe that $\vgedit{v}_{t+1}[r_1] = 0$ and $\vgedit{v}_{t+1}[r_2] = 1$, for any point $p$ outside this interval, it follows that
\begin{align*}
    \left|\hat{F}^{t+1} (p) - \frac{1}{t+1}\sum_{s=0}^{t+1} \vgedit{v}_s (p) \right|\
    &\leq \left|\frac{1}{t+1} \left(t \cdot \hat{F}^{t} (p) + \vgedit{v}_{t+1}(p) \right) - \frac{1}{t+1} \left( \sum_{s=0}^{t} \vgedit{v}_s (p) + \vgedit{v}_{t+1}(p) \right) \right|\ \\
    &\leq \left|\frac{1}{t+1} \left(t \cdot \hat{F}^{t} (p) - \sum_{s=0}^{t} \vgedit{v}_s (p) \right) \right|\ \\
\intertext{by inductive hypothesis}
    &\leq \frac{\log n - \log (p_2 - p_1)}{2(t+1)}
\end{align*}
Recall that the algorithm inserts a new point $r_3$ at the midpoint of $r_1,r_2$.
For a point $p$ in the interval $[r_1, r_2]$, 
\begin{align*}
    \left|\hat{F}^{t+1} (p) - \frac{1}{t+1}\sum_{s=0}^{t+1} \vgedit{v}_s (p) \right|\
    &\leq \left|\frac{1}{t+1} \left(t \cdot \hat{F}^{t} (p) + \frac{1}{2} \right) - \frac{1}{t+1} \left( \sum_{s=0}^{t} \vgedit{v}_s (p) + \vgedit{v}_{t+1}(p) \right)  \right|\ \\
    &\leq \left|\frac{1}{t+1} \left(t \cdot \hat{F}^{t} (p) - \sum_{s=0}^{t} \vgedit{v}_s (p) \right) + \frac{1}{t+1} \left(\frac{1}{2}   - \vgedit{v}_{t+1}(p) \right) \right|\ \\
\intertext{by inductive hypothesis}
    &= \frac{\log n - \log (r_2 - r_1)}{2(t+1)} + \frac{1}{2(t+1) } \\
    &= \frac{\log n - \log (r_2 - r_1) - \log 2}{2(t+1)} \\
    &= \frac{\log n - \log (r_2 - r_3)}{2(t+1)}
\end{align*}
\end{proof}

\end{document}